\documentclass[10pt,twocolumn,letterpaper]{article}

\usepackage{iccv}
\usepackage{times}
\usepackage{epsfig}
\usepackage{graphicx}
\usepackage{amsmath}
\usepackage{amssymb}
\usepackage{booktabs}
\usepackage{multirow}
\usepackage{enumitem}
\usepackage{xcolor}

\usepackage[breaklinks=true,bookmarks=false]{hyperref}
\usepackage[capitalize]{cleveref}
\crefname{section}{Sec.}{Secs.}
\Crefname{section}{Section}{Sections}
\Crefname{table}{Table}{Tables}
\crefname{table}{Tab.}{Tabs.}

\iccvfinalcopy 

\def\iccvPaperID{6057} 

\ificcvfinal\fi

\begin{document}

\title{Foreground-Background Separation through Concept Distillation from Generative Image Foundation Models}

\author{Mischa Dombrowski$^1$ \qquad Hadrien Reynaud$^2$ \qquad Matthew Baugh$^2$ \qquad Bernhard Kainz$^{1,2}$\\
$^1$Friedrich--Alexander--Universit\"at Erlangen--N\"urnberg \qquad
$^2$Imperial College London
\\
{\tt\small mischa.dombrowski@fau.de}
}

\maketitle
\ificcvfinal\thispagestyle{empty}\fi

\begin{abstract}
    Curating datasets for object segmentation is a difficult task. 
    With the advent of large-scale pre-trained generative models, conditional image generation has been given a significant boost in result quality and ease of use. 
    In this paper, we present a novel method that enables the generation of general foreground-background segmentation models from simple textual descriptions, without requiring segmentation labels. 
    We leverage and explore pre-trained latent diffusion models, to automatically generate weak segmentation masks for concepts and objects.
    The masks are then used to fine-tune the diffusion model on an inpainting task, which enables fine-grained removal of the object, while at the same time providing a synthetic foreground and background dataset. 
    We demonstrate that using this method beats previous methods in both discriminative and generative performance and closes the gap with fully supervised training while requiring no pixel-wise object labels. We show results on the task of segmenting four different objects (humans, dogs, cars, birds) and a use case scenario in medical image analysis. 
    The code is available at \href{https://github.com/MischaD/fobadiffusion}{https://github.com/MischaD/fobadiffusion}.
\end{abstract}

\section{Introduction}
\label{sec:intro}

Supervised pretraining, \emph{e.g.}, with ImageNet\cite{deng2009imagenet}, has demonstrated reduced training times and boosted performance. This gave rise to models that could be trained once over large amounts of data before being adapted to specialised tasks, such as image recognition, object detection, image segmentation\cite{simonyan2014very}, and medical image analysis\cite{schlemper2017deep}. 
The recent development of self-supervision techniques and their ability to learn without manual labels led to much larger scale training datasets \cite{changpinyo2021conceptual} and to the creation of foundation models \cite{bommasani2021opportunities}.
At present, the use of pre-trained models for a wide range of diverse downstream tasks defines a very active and intriguing area of research.

\begin{figure}
    \centering
    \includegraphics[width=\linewidth]{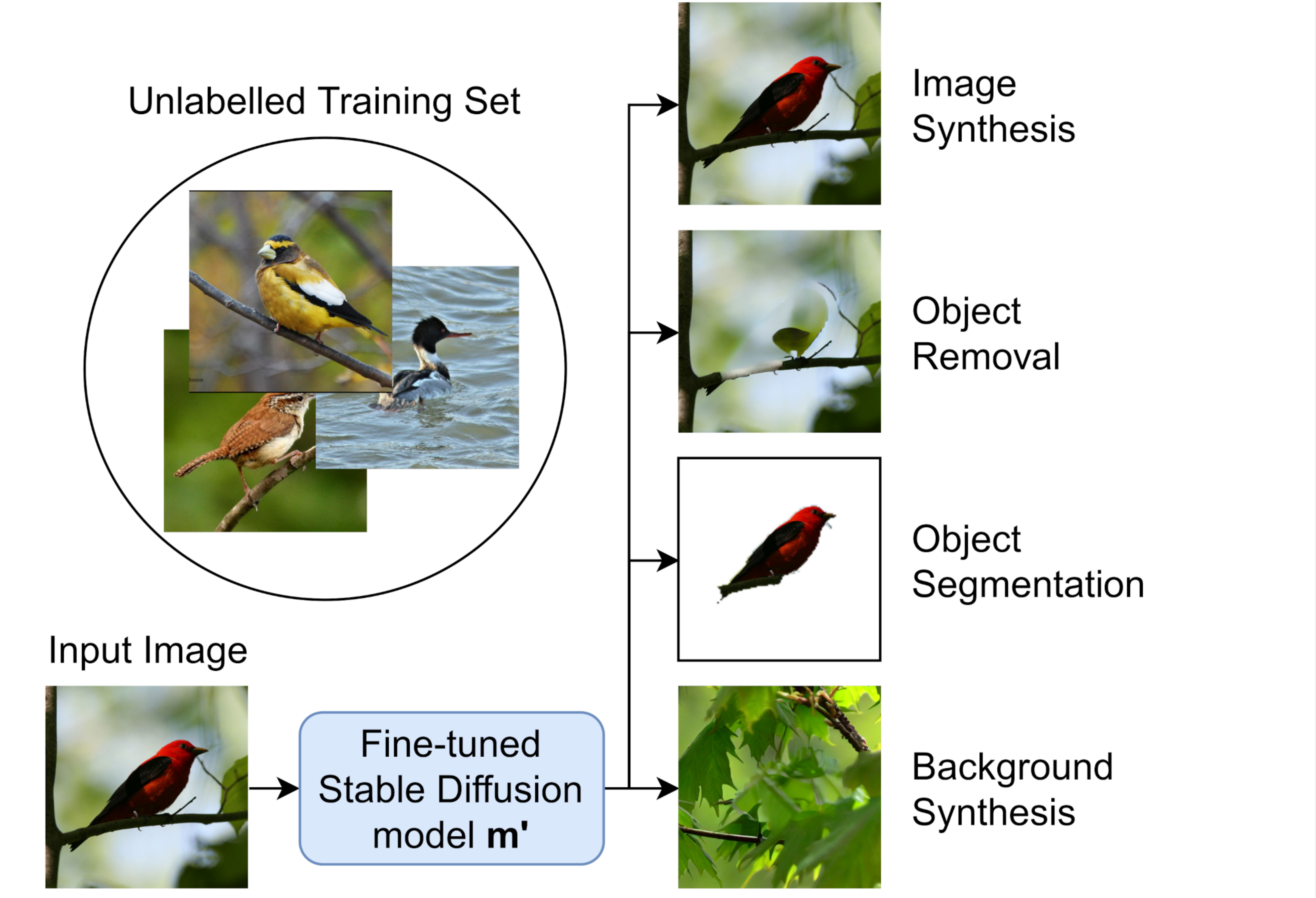}
    \caption{High level overview of our proposed method: Without needing a single labelled image, our method is able to generate foreground, background, and segmentation masks for any concept that is known to a text-to-image generative network.}
    \label{fig:model_abstract}
\end{figure}

Large scale foundation models are already established in natural language processing, with most of them being based on the Transformer architecture \cite{vaswani2017attention,devlin2018bert,brown2020language}. 
A crucial part of this architecture are cross- and self-attention layers, which compute interpretable importance weightings~\cite{parikhetal2016decomposable,cheng2016long}.

Diffusion models are based on the U-Net architecture \cite{ronneberger2015u} with additional attention layers~\cite{oktay2018attention} to condition on textual prompts. 
Therefore, we can extract inherently interpretable pixel importance scores from conditioning on textual prompts. 
Furthermore, the reverse diffusion process teaches the U-Net to successively remove noise from images, starting from pure Gaussian noise. In the early steps of this process, where the images resemble pure noise, the texture is non-existent, and the model only learns structures. 

Recently, latent diffusion models have emerged as state-of-the-art generative models for the task of text-to-image generation~\cite{dhariwal2021diffusion,rombach2022high_latent_stable_diffusion,ramesh2022hierarchical}.
However, training such models requires a significant amount of CO$_2$-intensive resources and until recently, pre-trained model weights have not been publicly available.
Rombach et al. were the first to published their weights and model architecture~\cite{rombach2022high_latent_stable_diffusion}, which facilitated the development of numerous derived applications~\cite{yang2022diffusion,croitoru2022diffusion,singer2022make} and established this model as the foundation model for tasks that require generalised representations of concepts in images.
State-of-the-art latent diffusion models are able to generate high resolution images of a vast amount of different objects, suggesting that a highly expressive latent representation of the data has been learned.

We hypothesize that we can leverage these learned latent representations for our own downstream tasks of zero-shot foreground-background generation.
Using a generative latent diffusion foundation model, we are able to extract a weak segmentation mask around an arbitrary object by computing the importance maps based on the textual input prompts. 
Weak segmentation masks have been shown to be an effective prior for segmentation models, given that enough training samples are available~\cite{lu2016learning,rajchl2016deepcut}.
We then use these preliminary masks to fine-tune a latent diffusion model on (1) generating new images from this dataset, as well as (2) inpainting regions where the object is not present according to our preliminary masks.
The resulting model is able to perform full-image synthesis, as well as foreground, background and mask generation, as summarized in \cref{fig:model_abstract}.
A segmentation model trained using these masks can then achieve a level of performance that is close to  direct supervision, despite not requiring manual segmentation masks at any point in the pipeline.
This suggests that labour-intensive ground-truth image annotation workflows could become obsolete in the future, and be replaced by concept distillation from generative foundation models.

\noindent Our main contributions are: 
\begin{itemize}[noitemsep]
    \item We propose a self-supervised, hyperparameter-free, approach for dataset-independent foreground-background segmentation, based on latent diffusion models, capable of synthesizing foreground, background, and segmentation masks.

    \item We describe a general framework to extract importance scores obtained from pretrained diffusion models and detail how to use them to improve segmentation performance. 

    \item We verify the feasibility of our method on a set of four different foreground background segmentation tasks, spanning humans, birds, dogs, and cars and show that our method achieves results close to supervised methods while being trained without direct supervision. 
    
    \item We experiment with the extension of our method to domain-adapted diffusion models by showing promising results on a medical segmentation task.
\end{itemize}


\section{Related Work}
\label{ref:related_work}

\noindent\textbf{Semantic Segmentation} refers to the identification of high level concepts in an image, which enable their extraction from the image. \cite{long2015fully,noh2015learning} introduced the use of fully convolutional networks for this task, which superseded previous shallow feature classification approaches~\cite{galleguillos2010context}.
Currently, the most common segmentation network architectures are designed as encoder-decoder pairs, as it enforces an information bottleneck that facilitates generalisation. The encoder provides meaningful low dimensional representations and the decoder reconstructs high-resolution segmentation maps~\cite{chen2017deeplab,ronneberger2015u,long2015fully,noh2015learning}.
Recent approaches maximise the use of multi-scale information with multi-scale attention~\cite{tao2020hierarchical}, squeeze-and-attention~\cite{zhong2020squeeze}, and Transformers~\cite{yuan2020object}. 
The final pixel classification operation, which creates the segmentation mask, is performed through multinomial logistic regression. These methods require large amounts of manually labelled training samples, which can be labour-intensive and expensive.

Weakly Supervised Semantic Segmentation partially mitigates that by learning from weak global labels, such as image-wise class labels, and perform rough semantic-segmentation tasks. 
These approaches often leverage the learned representations in intermediate layers through attention maps \cite{Jiang_2022_CVPR} or saliency maps \cite{lee2021railroad} extraction.
These representations are learned in a supervised manner through a classification task, as opposed to our approach that leverages an even weaker signal: free-form text-embeddings.
Furthermore, these methods cannot generate images and are, therefore, not suitable for foreground-background synthesis.

\noindent\textbf{Foreground background separation} is a segmentation task where the goal is to apply binary classification over all the pixels of an image to separate the object of interest, the foreground, from the contextual background. For example, in video analysis tasks, the background is often defined as parts of scenes that are at rest~\cite{sheikh2009background}. More recently, decomposing individual images into potential foreground and background layers became an intriguing research topic~\cite{ji2019invariant}. Since the foreground-background separation factors are not known a priori, many related works formulate the problem as a category-agnostic unsupervised segmentation approach.  
While deep neural networks can learn pixel clustering in an unsupervised way~\cite{ji2019invariant,kim2020unsupervised,ouali2020autoregressive}, it often leads to inferior performance when compared to fully supervised methods. Other works also tried to learn image segmentation from generative models, for example via direct sampling from the training distribution with cut-and-paste~\cite{remez2018learning}, image combinations via styleGAN~\cite{abdal2021labels4free}, erasing and redrawing~\cite{chen2019unsupervised}, and through inpainting~\cite{savarese2021information}. 

The work closest related to ours is~\cite{yang2022learningforeground_background_segm} who employ layered generative adversarial networks (GANs) to generate distinct images for the foreground and background. 
As opposed to \cite{yang2022learningforeground_background_segm}, we are not relying on the unknown structure of the latent manifold to separate the embedding codes that represent foreground and background, but instead follow the directly interpretable paradigm `textual concept description' $\rightarrow$ `segmentation model'. 
Both approaches introduce a similar bias, ours through the textual concept description and them through object-specific models, making our method similarly conditioned but more flexible.

\noindent\textbf{Diffusion models} are generative methods. 
Generative modeling has always been an important task in deep learning. Recently, Diffusion models gained a lot of attention thanks to the impressive results achieved by \cite{ramesh2021zero, saharia2022photorealistic}. Based on \cite{ho2020denoisingddpm, song2020denoisingddim}, these models currently define the state-of-the-art in conditional image generation, and have been extended towards  text-to-image models such as unCLIP \cite{ramesh2022hierarchical} and Stable Diffusion \cite{rombach2022high_latent_stable_diffusion}. 
Part of that success is due to the efficacy of straightforward extensions to diffusion models, such as classifier-free guidance \cite{ho2022classifier}. 
Current research about diffusion models focuses mostly on inpainting \cite{chung2022come,chung2022improving,rombach2022high_latent_stable_diffusion}. 
Depending on the task, it might be easier to learn the object itself and then train a diffusion model on the task of inpainting regions where the object is not present. 
However, we focus on the scenario where we have the object itself present and try to remove it. 

Shortly after latent diffusion models showed superiority in terms of image sample quality compared to GANs~\cite{dhariwal2021diffusion}, they were conditioned on the description of concepts~\cite{ramesh2021zero}.
These models became widely available online, including on commodity hardware after the computationally expensive denoising process was accelerated on a fundamental level~\cite{liu2022pseudo_plms}.
Direct translation of text into object-centric representations has been attempted but rather in the context of attention editing~\cite{hertz2022promptgoogleattentionediting}, subject specific image generation~\cite{ruiz2022dreambooth}, and textual concept-refined image-to-image translation~\cite{wang2022pretraining}.
Textual inversion, where examples of concepts are provided to a diffusion model as text and image tuples to teach the model a new concept~\cite{gal2022image}, operates on the same fundamental input level as our approach.
However, it is not able to provide object segmentation masks or concept-specific pixel importance scores. 
Recent advances on diffusion models have led to a spike in research around how latent features can be extracted \cite{hertz2022promptgoogleattentionediting,baranchuk2021label} however, their potential to be used for zero-shot segmentation has yet to be explored.




\section{Method}
\label{sec:prelimniaries}

\begin{figure*}
    \centering
    \includegraphics[width=\linewidth]{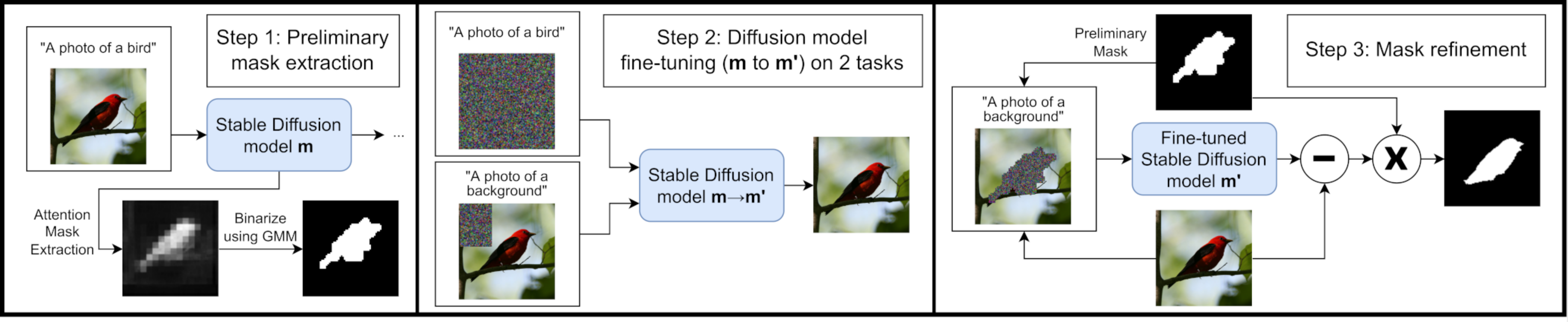}
    \caption{Overview of our model pipeline for self-supervised foreground-background segmentation.}
    \label{fig:ModelOverview}
\end{figure*}

Our approach is summarized in \cref{fig:ModelOverview}.
At a high level, we first use the attention maps from a pretrained latent diffusion model to compute coarse foreground segmentations (preliminary masks).
We use these masks to fine-tune the diffusion model to be able to remove the object from the image, replacing it with background information, whilst also being able to generate new samples from the true distribution $p(\mathbf{x})$ of the dataset.
By comparing the original images to those with the background inpainted over the foreground, we are then able to produce much more refined masks.
We can also then use the fine-tuned model to sample an arbitrary number of new images, including only the foreground, only the background, and the masks separating them.
The only assumptions this method uses is that we have access to a dataset of images where the chosen object is always present, and that the foundation model we use has learned a meaningful representation of the object we want to identify.
In this context, objects can describe concepts like birds, cars, dogs, humans, but also parts of objects such as arms or legs.

We start from the latent diffusion model (LDM) \cite{rombach2022high_latent_stable_diffusion}, a foundation model pretrained on the task of text-to-image generation, denoted as $m$.
It generates images by sampling gaussian noise and iteratively denoising them in $T$ diffusion steps. Throughout this work, we keep the default value from \cite{rombach2022high_latent_stable_diffusion} of $T=50$ steps.

Let \textbf{D} denote a dataset where all the images contain our desired object.
Formally, we aim to generate unsupervised segmentation masks of the original dataset $\mathbf{D_m}$ as well as  a synthetic dataset that contains synthetic images $\mathbf{D'}$, masks $\mathbf{D'_m}$, foreground $\mathbf{D'_f}$, and background $\mathbf{D'_b}$ denoted as set 
$\mathbf{D'_s} = \{ \mathbf{D'}, \mathbf{D'_m}, \mathbf{D'_f}, \mathbf{D'_b}\}$. 

\noindent\textbf{Preliminary Masks:}
\label{subsec:preliminary_mask_meth}
The LDM was trained on paired text-image data. Thus, the output is conditioned on a text input, which we have to carefully choose to generate our initial attention maps. 
In practice, self- and cross-attention work well for conditioning on different inputs, especially across different modalities \cite{vaswani2017attention}. 
As input, we propose the prompt ``\textit{a photo of a \{object\}}" where ``object" is a high level description of our foreground object (\emph{e.g.} ``bird").


The first step in computing the preliminary masks is to leverage the raw attention maps computed in every cross-attention layer of the U-Net architecture.
We decide against cherry-picking different attention layers for different tasks to remain task-agnostic, but would like to point out that this could lead to task-specific improvements as an hyper-parameter option.  

Let $z_0$ denote the latent space representation of some input image x. In each layer, attention is computed as:

\begin{equation}
    \text{Attention}(Q, K, V) = \psi(Q_l, K_l^T) \cdot V_l
\end{equation}

\noindent with the attention probabilities $\psi(Q_l, K_l^T)$ defined as: 

\begin{equation}
    \psi_{\mathbf{z_{t,l}}}(Q_l, K_l^T) = \text{softmax}(\frac{QK^T}{\sqrt{d}})
\end{equation}

with $Q = W_Q^{(i,l)} \cdot \phi(\mathbf{z_{t,l}})$, $K = W_K^{(i,l)} \cdot \tau_{\theta}(y)$, and $V = W_V^{(i,l)} \cdot \tau_{\theta}(y)$
denoting the learnable projection matrices according to \cite{vaswani2017attention}, 
$\phi(\mathbf{z_{t,l}})$ the latent code of the l-th U-Net layer in the t-th reverse diffusion step, and $\tau_\theta(y)$ the learned latent representation for the textual input prompt \cite{rombach2022high_latent_stable_diffusion}.
$z_{t,l}$ denotes the latent representation of the stable diffusion model in layer l, conditioned at diffusion t.

The next step is to compute the mean attention maps $\hat{M}$ as expectations over multiple repeated diffusion steps according to 

\begin{equation}
\label{eq:eq_last_diff_step}
    \hat{M} = \sum_{t=1}^{T_0}\mathbb{E}_{\mathbf{z} \sim p(\mathbf{z_t}|\mathbf{z_{t+1}}, \mathbf{z_0})}
    [
    \sum_{l} \psi_{\mathbf{z_{t,l}}}(Q_l, K^T_l)
    ].
\end{equation}

We empirically show in our supplementary material that this can be simplified to performing single reverse diffusion steps which leads to a simplified importance score

\begin{equation}
    \label{eq:multiple_diff_steps}
    \hat{M} = \sum_{t=1}^{T_0}\sum_{l}{\psi_{\mathbf{z_{t,l}}}(Q_l, K_l^T)}
\end{equation}

\begin{figure}[t]
    \centering
    \includegraphics[width=\linewidth]{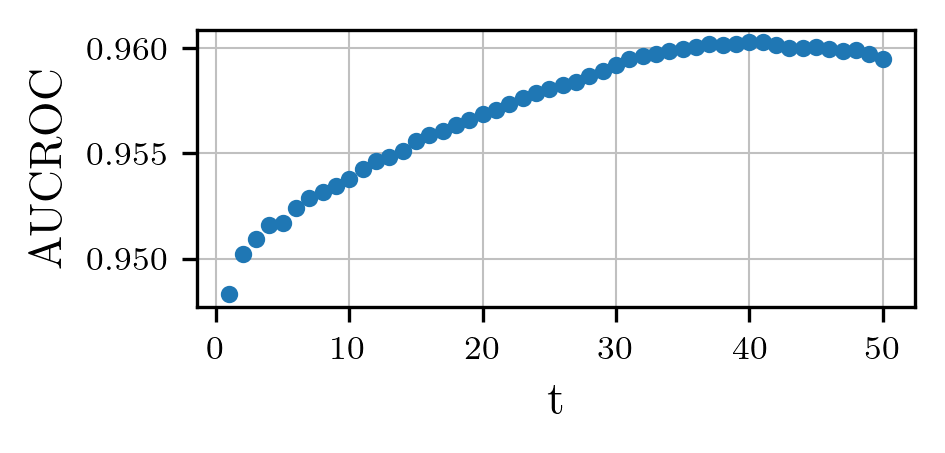}
    \caption{$T_0$ vs AUC-ROC on CUB. Incorporating more reverse diffusion steps into the attention computation improves the AUC-ROC against the groundtruth only up to roughly $T_0 = 40$. }
    \label{fig:rev_steps_vs_auc}
\end{figure}

\Cref{fig:rev_steps_vs_auc} shows AUCROC of using $\hat{M}$ to identify the foreground as a function of the number of reverse steps. 
There is a clear improvement visible computing the mean over up to roughly $T_0 = 40$ reverse diffusion steps. 
Including higher diffusion steps seems to deteriorate the accuracy of $\phi$. This is because for higher values of $t$ the input images approach pure Gaussian noise. 
However, for medium  values of $t$ the input and output images already approximate the basic structure of the final objects. Other details, like texture, appear at later stages but do not bear any valuable information for this use case.  
This observation could mean that for time critical applications, starting from lower $t$ should suffice.

The third and final step in retrieving the preliminary masks is to binarize the mean attention maps. To do so, we take advantage of the observation that all attention maps resemble a bimodal distribution, with one mode at a low value for the non-object pixels and one mode at a high value for the object pixels.
Hence, we model absolute values of the instance-wise attention scores as a bimodal Gaussian mixture model (GMM) to produce GMM masks. Additionally, we remove orphan pixels by computing the mean filter over the resulting binary classification map to produce our preliminary masks $M_{pre}$. 
These preliminary masks could potentially already be used to detect objects as evaluated in \cref{sec:evaluation}. 

\noindent\textbf{Fine-tuning and Mask Refinement:}\label{subsec:refined_masks}
The main problem with the preliminary masks is that, because they are derived from the latent space of the LDM, their resolution is only $64 \times 64$.
This limits them to being a coarse segmentation, which often overestimates the size of the object rather than following the object's sharp edges.
A rough segmentation is sufficient for tasks like inpainting, especially if the area around the object is homogeneous, but for the task of extracting the foreground object it produces unwanted artefacts. 
To work around this, we leverage the same diffusion model that we used to extract the masks. 
We use the binary classification prediction of the GMM mask $M_{pre}$ to select random rectangles within the image that only contain background.
Then we fine-tune the model to inpaint only the background of these images by conditioning on the prompt $y_b = $``a photo of a background" using the image as ground truth (See supplements for examples).
Simultaneously, the model is fine-tuned on the task of full image synthesis to generate new samples for $\mathbf{D'}$ by conditioning on $y_f =$ ``a photo of a \{object\}".

To generate the final foreground masks $\mathbf{D_m}$  we use the fine-tuned LDM $m'$ to inpaint the background over the area covered by the preliminary mask conditioning on the background prompt.
To identify the foreground region, we take the pixel-wise intensity difference between the background-inpainted and original images.
As the inpainted images are conditioned to generate background, the difference is higher in the true foreground region.
We apply a Gaussian mixture model to the pixel-wise difference map to create a binary classification map.
Formally the refined masks $M$ are computed as 

\begin{align}
\label{eq:refined}
    M = M_{pre,up} \odot g(| \mathbf{x} - m'(\tilde{\mathbf{z_0}}, y_b) |),\\
    \tilde{\mathbf{z_0}} = \mathbf{z_0} \odot (1 - M_{pre}) + \mathbf{z} \odot M_{pre}
\end{align}

where $M_{pre,up}$ denotes the preliminary mask upsampled to the input image's resolution, $g$ applies the bimodal Gaussian mixture model to the pixel values and $\tilde{\mathbf{z_0}}$ is the latent code of an image with the foreground region replaced with random noise $\mathbf{z}$. 
This produces refined masks, as the use of pixel-wise error improves the segmentation around the sharp edges of the object.
Additionally, the computation of the refined masks is performed in pixel-space of the images instead of the latent space of $m$ and therefore produces even more detailed masks.

Finally, to further improve our mask prediction we use the refined masks as labels to train a U-Net \cite{ronneberger2015u} to directly segment the foreground of the images, similar to the approach in \cite{yang2022learningforeground_background_segm}, which follows a standardized approach of training a U-Net on a fixed number of steps and hence does not require any hyperparameter tuning.
We experiment with training the segmentation network on refined masks of the original unlabelled training data ($\mathbf{D_s}$), as well as training it with the fully synthetic dataset $\mathbf{D'_s}$ as an augmentation method, which we generate by prompting the fine-tuned model with the foreground conditioning $y_f$ and repeating our pipeline of mask refinement on this synthetic dataset to get segmentation labels.

\section{Evaluation and Results}
\label{sec:evaluation}

\noindent\textbf{Implementation:}\label{subsec:implementation}
We use PyTorch 1.11 and run our experiments on a workstation with two A6000 Nvidia GPUs. Concept distillation training takes on average one day. The forward pass is fast, equivalent to that of a standard U-Net. 

\noindent\textbf{Datasets:}
We choose our datasets such that they cover a variety of different objects, including in-the-wild animals and cars, as well as humans in static setups.

Human3.6m \cite{h36m_pami_human3point6} is a dataset of 3.6 million images of humans in different scenarios and situations. To show that our method does not rely on large datasets we take a subset of 6000 randomly chosen images centred around the human and cropped to $256 \times 256$ pixels from the training dataset of Human3.6m.

To test the method on representations of animals, we use two datasets:
the Stanford Dog Dataset \cite{KhoslaYaoJayadevaprakashFeiFei_FGVC2011}, which contains 20,580 images of dogs divided into different categories, and the Caltech-UCSD Birds 200 (CUB) dataset \cite{WahCUB_200_2011}, which contains 11,788 images of birds from 200 different species.

Finally, we also experiment with the detection of cars using \cite{KrauseStarkDengFei-Fei_3DRR2013_stanford_cars}, which consists of 16,185 images of cars in different natural and non-natural settings. 
All these datasets come with subcategories grouping the images based on selected features. For our use case, we consolidate these groups when prompting the model and use the classes \textit{cars}, \textit{dogs}, \textit{human}, and \textit{birds} to simulate the absence of manual labels.

\begin{table}[ht]
    \centering
            \begin{tabular}{lccc}
            \toprule
            \multirow{2}{*}{Methods} & \multicolumn{3}{c}{ CUB } \\
            & ACC & IoU & mIoU \\
            \midrule Fully supervised U-Net & $97.9$ & $88.3$ & $93.0$ \\
            \midrule GrabCut \cite{rother2004grabcut} by \cite{savarese2021information} & $72.3$ & $36.0$ & $52.3$ \\
            ReDO \cite{chen2019unsupervised} & $84.5$ & $42.6$ & $-$ \\
            PerturbGAN \cite{bielski2019emergence} & $-$ & $-$ & $38.0$ \\
            IEM + SegNet \cite{savarese2021information}& $89.3$ & $55.1$ & $71.4$ \\
            Melas-Kyriazi et al. \cite{melas2021finding} & $92.1$ & $66.4$ & $-$ \\
            Layered GAN \cite{yang2022learningforeground_background_segm} & $9 4 . 3$ & $6 9 . 7$ & $8 1 . 7$ \\
            \midrule
            Ours (U-Net trained on $\mathbf{D_s}$) & $95.2$ & $75.1$ & $84.8$ \\
            Ours (U-Net trained on $\mathbf{D_s} \cup \mathbf{D'_s}$) & $\mathbf{95.6}$ & $\mathbf{77.2}$ & $\mathbf{86.0}$ \\
            \bottomrule
            \end{tabular}   
    \caption{Comparison to other segmentation methods. Baselines taken from \cite{yang2022learningforeground_background_segm}. 
    Details on the training of the U-Nets can be found in the appendix.}
    \label{tab:comparison_to_layers_and_supervised}
\end{table}

\begin{figure*}[ht]
    \centering
    \includegraphics[width=0.98\linewidth]{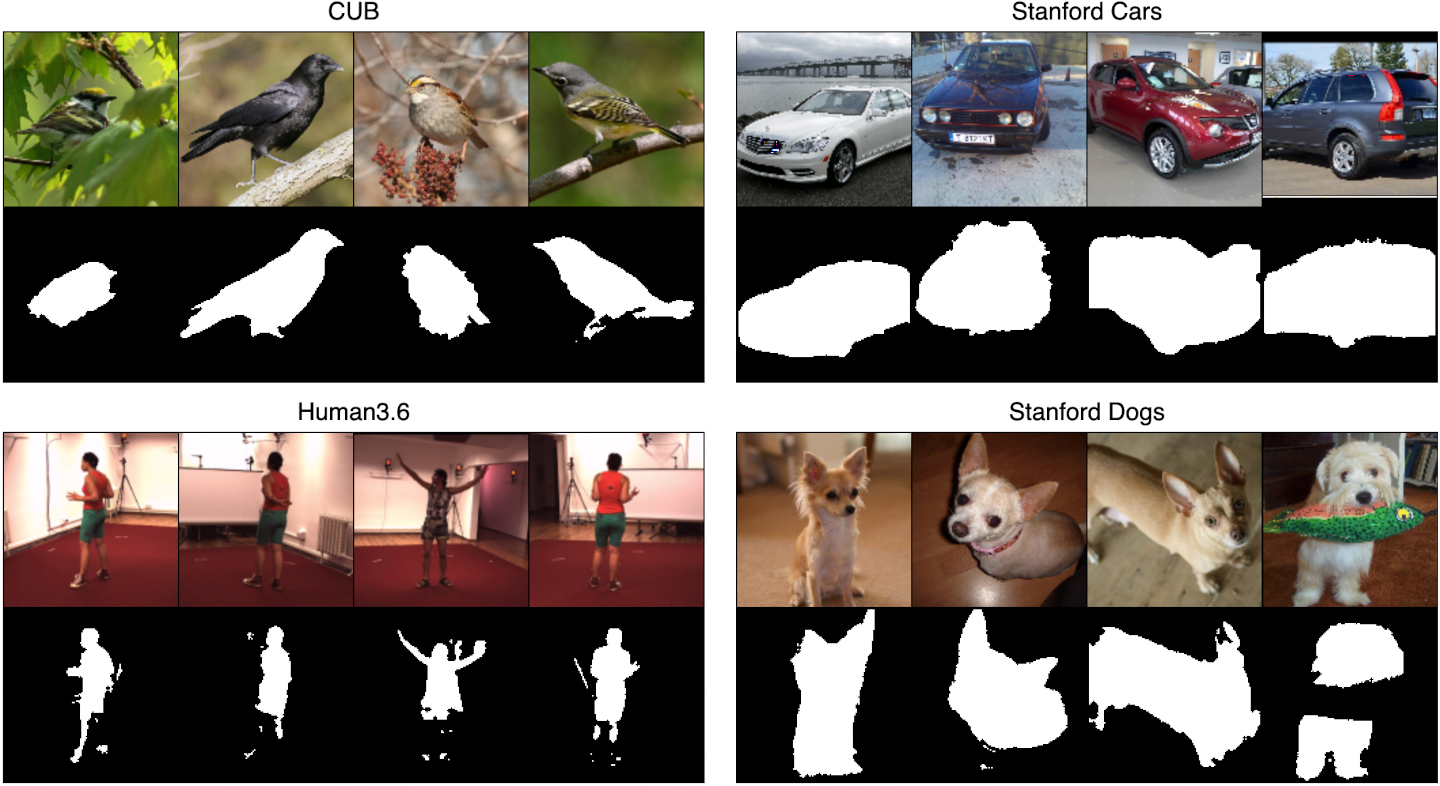}
    \caption{Unsupervised Segmentation Masks generated by our proposed approach. }
    \label{fig:final_segmenation_masks}
\end{figure*}
\noindent\textbf{Self-supervised Segmentation Performance:}
\Cref{tab:comparison_to_layers_and_supervised} compares our methods performance against other unsupervised methods on the CUB dataset, showing the pixel-wise accuracy, the Intersection over Union (IoU) of the foreground segmentation and the mean IoU.
From this we see that training a U-Net on our self-supervised labels produces a model that outperforms all other methods, achieving an overall foreground IoU improvement of 5.4 compared to \cite{yang2022learningforeground_background_segm}.
Furthermore, by adding the fully synthetic dataset $\mathbf{D'_s}$ to the training data we are able to improve the performance even further, reaching a foreground IoU of 77.2. 
\Cref{tab:comparison_to_layered_fidnmiiou} also shows high foreground IoU values across the other datasets, with \cref{fig:final_segmenation_masks} showing qualitative examples.

\begin{figure*}[ht]
    \centering
    \includegraphics[width=0.98\linewidth]{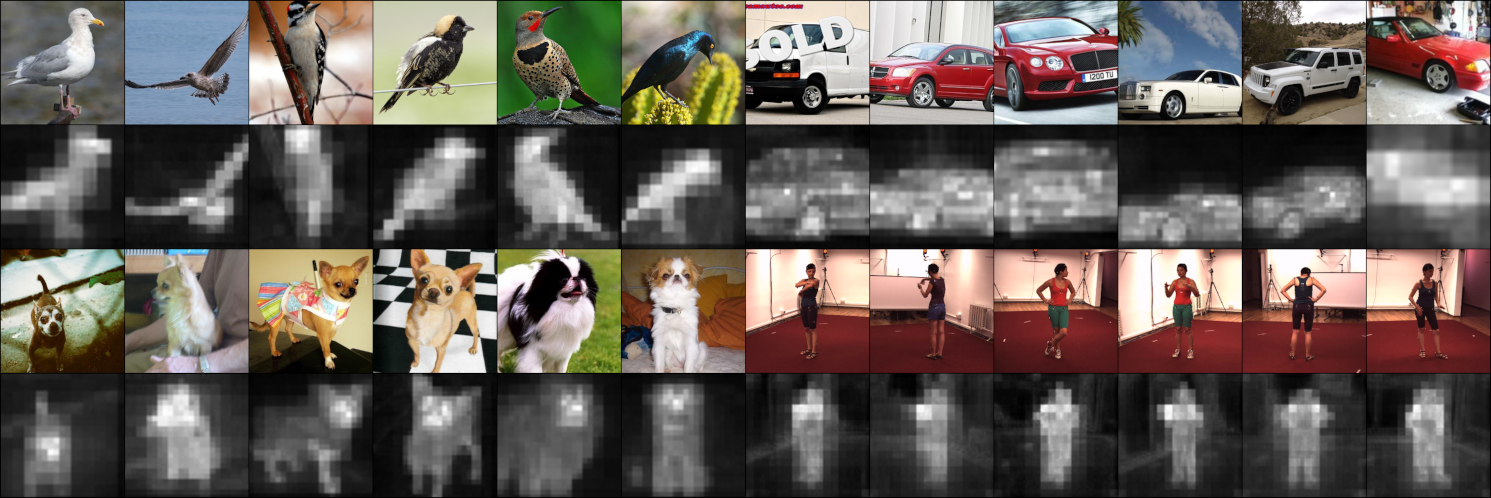}
    \caption{Mean attention maps for all datasets in latent space \textbf{z} of the diffusion model. Prompts are ``a photo of a \{object\}", where \{object\} is replaced by ``bird" for the first pair of rows, then ``car", ``dog", and ``human".}
    \label{fig:prelim_masks_all_datasets}
\end{figure*}

\noindent\textbf{Mean Attention Map Performance:}\label{subsec:att_map_perf}
Computing the mean attention maps $\hat{M}$ and comparing them to the ground-truth yields a remarkable AUC-ROC for the bird dataset of 97.1. 
Scores are normalized instance-wise to a range of 0 to 1. We also experiment with no normalization, which gives a slightly worse AUC-ROC of 97.08.
Qualitative examples from all datasets are shown in \cref{fig:prelim_masks_all_datasets}, displaying the mean attention maps' ability to localise the foreground in different scenarios.
To compare this to our classification results we compute the threshold such that we reach over 95\% true positive rate on a reserved training set of 100 images.
True positive rate is more important in our case because we observe that falsely classifying pixels as background leads the inpainting model taking these foreground pixels as sources to inpaint larger parts of the image. 
We reach a pixel-wise accuracy of 86\% on a set of 1000 test images suggesting that our method of extracting the classification masks already can provide meaningful results. 

While these results are encouraging, they also require a ground-truth dataset and thresholding that we do not want to rely on.
The results indicate that our computed masks are very good at locating the objects, albeit they do not reach state-of-the-art performance despite adding supervision.
Training on refined masks and integrating synthetic data surpasses supervised results without using any labels as shown in the next section. 

\begin{figure}[t]
    \centering
    \includegraphics[width=\linewidth]{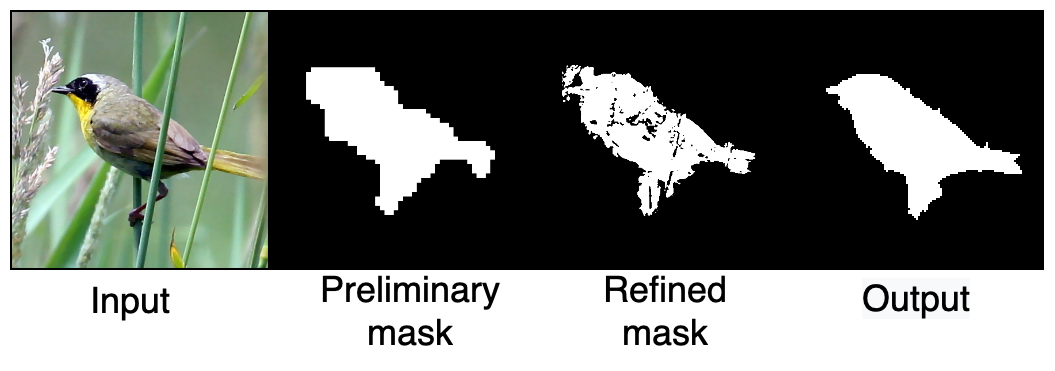}
    \caption{Progressive refinement of the segmentation masks.}
    \label{fig:mask_progression}
\end{figure}
\begin{table}[ht]
    \centering
            \begin{tabular}{lccc}
            \toprule \multirow{2}{*}{Methods} & \multicolumn{3}{c}{ CUB } \\
            & ACC & IoU & mIoU \\
            \midrule
            Preliminary Masks $M_{pre}$ & $83.5$ & $29.0$ & $55.7$ \\
            Simple Inpainting $M_{crop}$ & $90.0$ & $30.7$ &  $60.0$\\
            U-Net trained on ($\mathbf{D}$, $M_{pre}$) & $91.7$ & $66.5$ & $78.2$ \\
            Refined Masks $M$ & $92.4$ & $63.6$ & $77.4$ \\
           
            U-Net trained on $\mathbf{D_s}$ & $95.2$ & $75.1$ & $84.8$ \\
            U-Net trained on $\mathbf{D_s} \cup \mathbf{D'_s}$ & $\mathbf{95.6}$ & $\mathbf{77.2}$ & $\mathbf{86.0}$ \\
            \bottomrule
            \end{tabular}   
    \caption{Segmentation results for different steps of our pipeline.}
    \label{tab:ablationstudies}
\end{table}

\noindent\textbf{Ablation study:}
\label{subsec:intermediate_results}
\Cref{tab:ablationstudies} shows an ablation study for the individual components of our method. 
Initially, the zero-shot preliminary masks generated from the foundation model achieve good accuracy (83.5), but poor foreground IoU (29.0).
This reflects our earlier intuition that masks overestimate the size of the object due to the maps being computed at a lower resolution. 
Thanks to this, the GMM overestimates the boundaries and spares the need for any hyperparameter. 
Optimizing the threshold to maximise the accuracy over a set of 100 training images would increase the accuracy to 93.9\%, but requires a ground-truth dataset. 

Training a U-Net on these labels ($M_{pre}$) increases the IoU to 66.5, showing the value in training the segmentation network.
Using the refined masks as foreground segmentation gives comparable performance, with higher accuracy but lower foreground IoU.
Additionally, we experiment with replacing the inpainting step used to improve $M_{pre}$ with a simpler approach that crops background areas and uses them as inpainting in \cref{eq:refined} instead of $m'$. 
The resulting masks $M_{crop}$ are worse than the masks from our proposed refinement step (For more details see supplements). 
However, training a U-Net on these refined masks produces the best results, being further boosted by incorporating additional synthetic data.
The progressive refinement of the segmentation masks is  shown in \cref{fig:mask_progression}. 

\begin{table*}[ht]
\centering

\begin{tabular}{lc cc cc cc cc} 
\toprule
\multirow{2}{*}{Methods} & \multirow{2}{*}{Sup.} & \multicolumn{2}{c }{CUB}           & \multicolumn{2}{c }{Stanford Dogs} & \multicolumn{2}{c}{Stanford Cars}   & \multicolumn{2}{c}{Human3.6m}   \\
                         &                       & FID $\downarrow$ & IoU $\uparrow$  & FID $\downarrow$ & IoU $\uparrow$  & FID $\downarrow$ & IoU $\uparrow$   & FID $\downarrow$ & IoU $\uparrow$   \\ 
\midrule
FineGAN           \cite{singh2019finegan}       & Weak                  & $23.0$           & $44.5$          & $54.9$           & -               & $24.8$           & -                & -               & -               \\
OneGAN          \cite{benny2020onegan}          & Weak                  & $20.5$           & $55.5$          & $48.7$           & -               & $24.2$           & -                & -               & -               \\
LayeredGAN   \cite{yang2022learningforeground_background_segm}           & Unsup.                & $12.9$           & $69.7$          & $59.3$           & -               & $19.0$           & -                & -               & -               \\ 
\midrule
Ours                     & Self                  & $\mathbf{9.8}$   & $\mathbf{75.1}$ & $\mathbf{43.1}$  & $\mathbf{63.8}$ & $\mathbf{13.4}$  & $\mathbf{55.2}$  & \textbf{63.7}               &  $\mathbf{69.2}$              \\
\bottomrule
\end{tabular}
\caption{Quantitative Results on $\mathbf{D_s}$. Training details are shown in the supplements. Values are taken from \cite{yang2022learningforeground_background_segm}. Source code for \cite{singh2019finegan,benny2020onegan,yang2022learningforeground_background_segm} was not available for re-evaluation on the Dogs, Cars and Human3.6m datasets. 
IoU on CUB are reported using the prediction of our model and the ground truth  provided with the dataset. For the other datsets we use the IoU of the bounding boxes.  }
\label{tab:comparison_to_layered_fidnmiiou}
\end{table*}

\noindent\textbf{Data synthesis:}
\Cref{tab:comparison_to_layered_fidnmiiou} compares the generative ability of our fine-tuned diffusion model using the Fréchet inception distance (FID).
Our method achieves higher generation quality than all other methods across the CUB, Stanford Dogs, and Stanford Cars datasets.
We improve upon LayeredGAN's remarkably low FID scores for CUB and Stanford Cars by 3.1 and 5.6 respectively.

\begin{figure*}[ht]
    \centering
    \includegraphics[width=\linewidth]{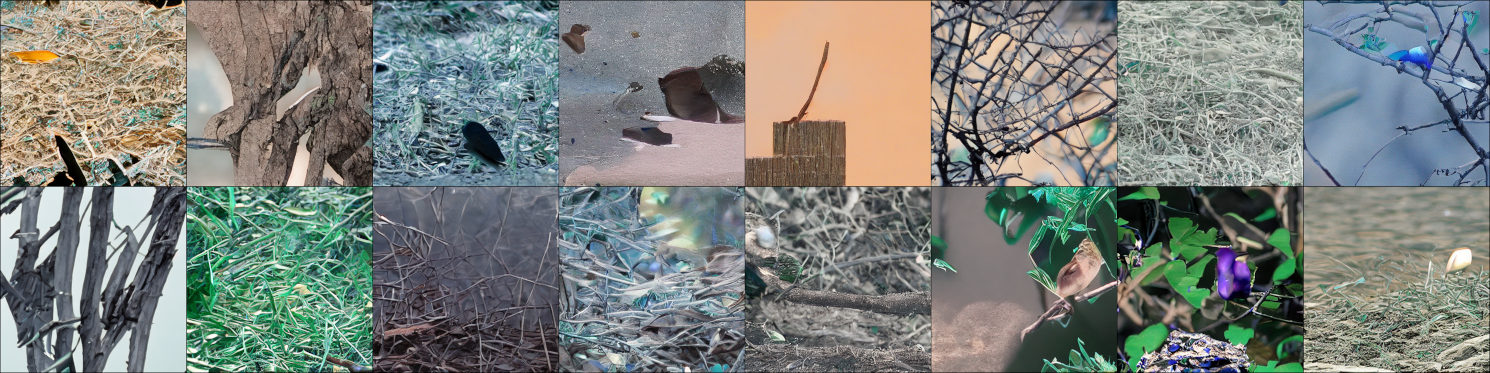}
    \caption{Full background image synthesis from the fine-tuned model, conditioned on $y_b$. Using our proposed fine-tuning method, the diffusion model is successfully able to generate images without birds from a dataset only consisting of images with birds. }
    \label{fig:full_image_synthesis}
\end{figure*}

Our method allows for the generation of samples covered entirely by the background, as shown in \cref{fig:full_image_synthesis}, without ever seeing such an image during training.
The success of this component is what enables the refined masks to be generated, as accurately inpainting the foreground allows us to use the pixel-wise difference between the original and inpainted images to precisely identify the foreground.

\begin{figure}[ht]
    \centering
    \includegraphics[width=\linewidth]{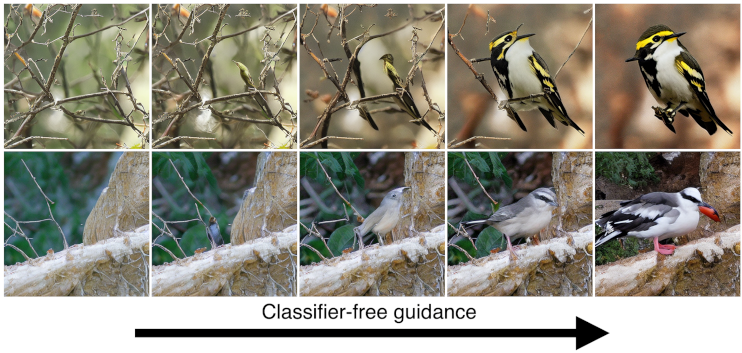}
    \caption{Synthetic results of $m'$ with changing scales of classifier-free-guidance, ranging from -2 on the left to +7.5 on the right.}
    \label{fig:concept-distill}
    \hfill
\end{figure}

\noindent\textbf{Concept Distillation:}
Finally we can evaluate if our model has indeed learned to distinguish between foreground and background by looking at the output of different classifier-free guidance scales starting from the same seed $\mathbf{z_t}$. 
Since we have fine-tuned our model only on two distinct textual prompts the conditional image generation should have collapsed to two clusters, one for the object and the other one for the background. Hence, instead of performing classifier-free guidance using the predictions of $m'$ conditioned on empty prompts we can directly use the predictions of $m'(\mathbf{z_t}, y_b)$ and $m'(\mathbf{z_t}, y_f)$ to perform image interpolation. 
We define classifier-free guidance in the direction of the foreground, hence, a higher score means that latent representations are pushed further in the direction of the object. 
The results are shown in \cref{fig:concept-distill}. For negative guidance scales, the background is more detailed
and there is no bird present. Increasing this scale leads to less detail in the background while birds often seem to naturally grow from the details in the background.
The quality of the birds visually improves while the quality of the rest of the image keeps degrading resulting in less detailed backgrounds.
We confirm this quantitatively by computing the FID for different classifier-free guidance scales. Without classifier-free guidance, \emph{i.e.}, scale = 1, the method reaches a FID of 9.8, at scale = 3 a FID of 11.3, and at scale = 7.5 a FID of 22.3.

\begin{figure}
    \centering
    \includegraphics[width=\linewidth]{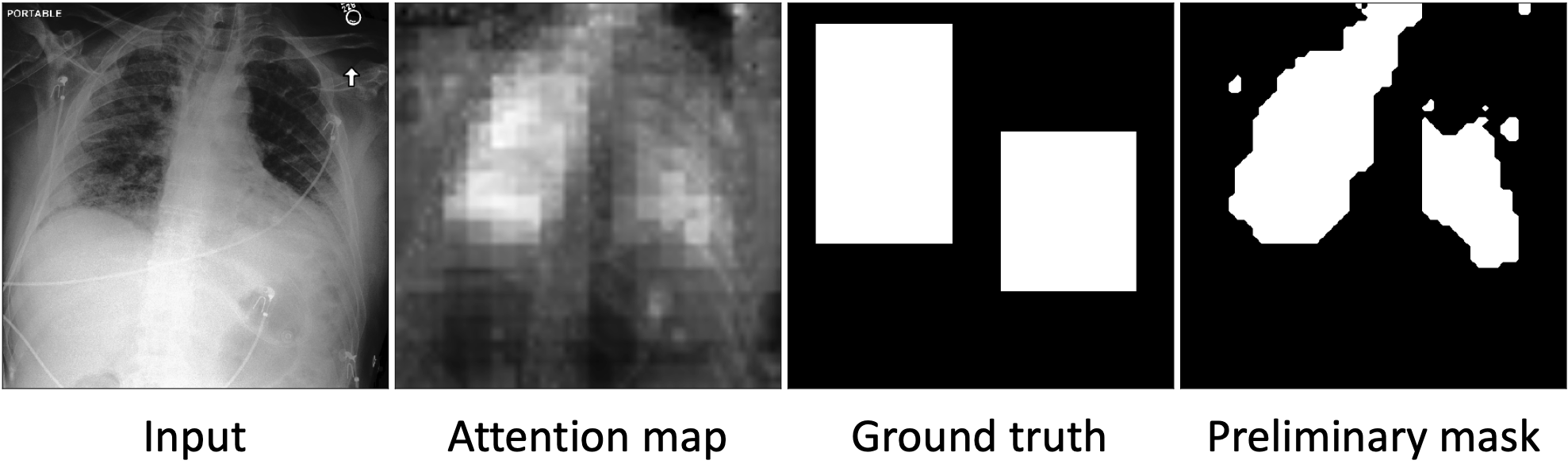}
    \caption{$\hat{M}$ and $M_{pre}$ extraction on a medical task.}
    \label{fig:medical_image_analysis}
\end{figure}

\noindent\textbf{Medical Image Analysis:} 
We want to evaluate if this approach can be applied to other domains, such as medical imaging. 
Since the LDM does not have any medical understanding, we first need to fine-tune it using MIMIC \cite{johnson2019mimic}, which provides chest x-ray images paired with radiology reports. We can fine-tune the model using a similar approach as the one suggested by \cite{chambon2022roentgen} (Details on the fine-tuning can be found in the supplements). 
Then we report the pixel-wise AUC-ROC on MS-CXR \cite{boecking2022making}, a subset of MIMIC with bounding box labels for diseases. 
Qualitative results can be seen in \cref{fig:medical_image_analysis}. 
The pixel-wise accuracy of the attention mask is already at 79.6\% AUC-ROC across eight different diseases, however, the bimodal GMM assumption no longer holds in many cases because the model distinguishes three regions, namely: background, foreground, and the rest of the chest region.

\section{Discussion}

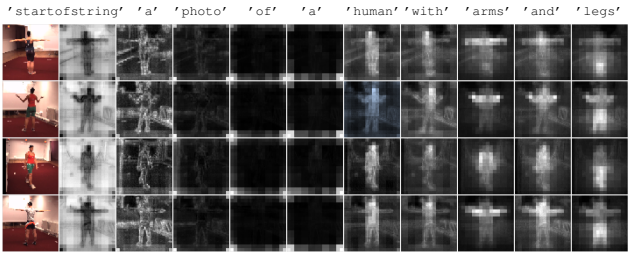
\begin{figure}[ht]
    \centering
    \begin{picture}(1000,100)
    \put(0,0){\includegraphics[width=\linewidth]{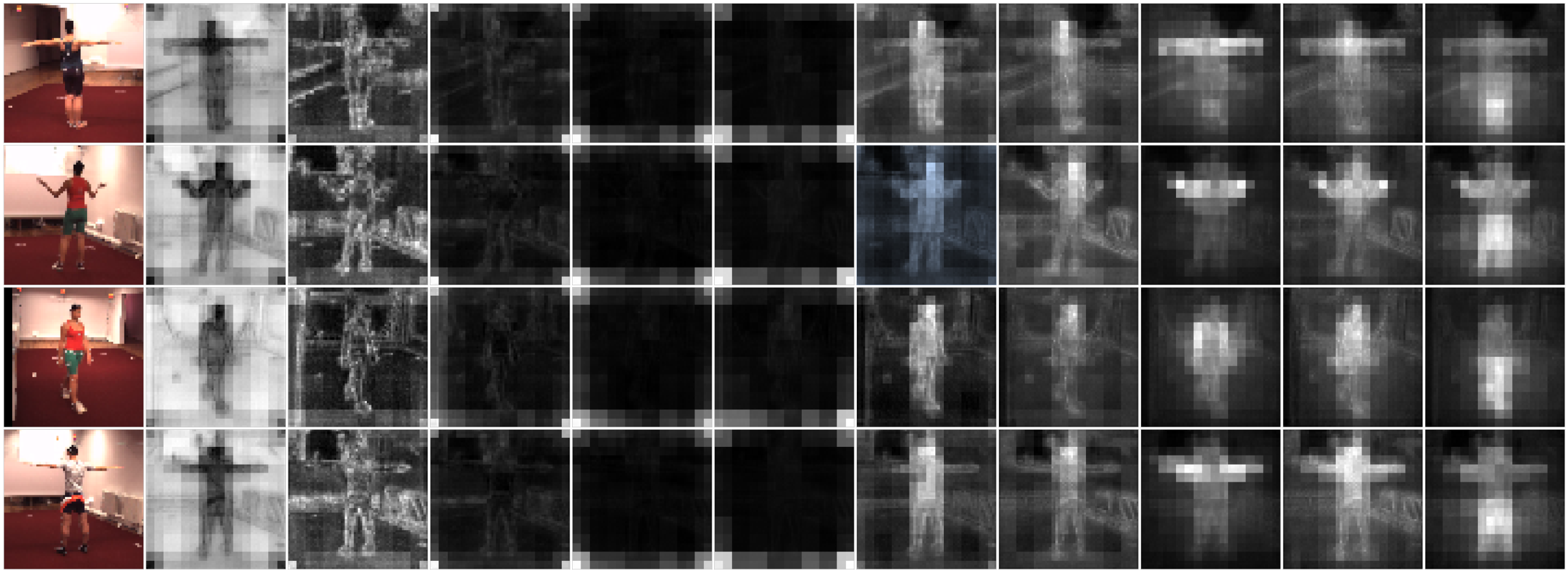}}
    \put(2,90){\tiny{\texttt{'startofstring'}}}
    \put(51,90){\tiny{\texttt{'a'}}}
    \put(65,90){\tiny{\texttt{'photo'}}}
    \put(93,90){\tiny{\texttt{'of'}}}
    \put(113,90){\tiny{\texttt{'a'}}}
    \put(130,90){\tiny{\texttt{'human'}}}
    \put(152,90){\tiny{\texttt{'with'}}}
    \put(175,90){\tiny{\texttt{'arms'}}}
    \put(197,90){\tiny{\texttt{'and'}}}
    \put(217,90){\tiny{\texttt{'legs'}}}
    \end{picture}
    \caption{Input image and mean attention maps for every word of the textual input prompt ``startofstring a photo of a human with arms and legs". The attention of ``human" focuses on the torso, the one for the ``arms" on the arms, and the one for the ``legs" on the legs. }
    \label{fig:attention_human_arms_legs}
\end{figure}

We show in \cref{sec:evaluation} that our method to extract the segmentation masks $M$ 
from the preliminary attention masks yields better results than computing an optimised threshold over a reserved miniset. 
This is possible because comparing to the inpainted background provides sharp edges around the object. 
However, our model is currently limited to detecting single object concepts.
An extension to multiple objects could be achieved by prompt engineering in combination with data augmentation techniques.
Taking the foreground masks and using them to extract objects would enable multi-instance and multi-object segmentation by layering multiple objects over each other and extending the final U-Net to a multi-label segmentation model.

Furthermore, learning from weak labels has the disadvantage that the segmentation model may learn and reproduce weaknesses of the initial method.
In our case, the bimodal GMM fails if the image has more than two distinct contrast clusters.
This is especially the case on the Human3.6m dataset where the floor, walls, and person have vastly different contrast-levels.
Consequently, the final segmentation sometimes fails to detect the lower part of the body as shown in \cref{fig:final_segmenation_masks}. 
However, our method could easily be adapted to only a part of the human body, such as the legs. 
We show this in \cref{fig:attention_human_arms_legs}, as by conditioning images from~\cite{h36m_pami_human3point6} on the prompt ``\textit{A photo of a human with arms and legs}" and computing $\hat{M}$ for the three concepts  (human, arms, and legs) we are able to produce attention maps focused on specific body parts.

\section{Conclusion}

In this work, we have presented a generalizable framework to train segmentation networks without any hyperparameter tuning using an unsupervised zero-shot approach following the paradigm of `textual concept description` $\rightarrow$ `segmentation model`.
We leverage the power of large generative latent diffusion models and fine-tune the model on the task of generating foreground and background images, which can be used as data augmentation methods. 
We show, that this method can achieve results close to supervised methods, without requiring any manually generated ground truth labels. 
Our approach is amenable to supervised deep learning and can be combined with existing models to boost segmentation performance even further. 

In future work we will explore how multi-object, multi-instance segmentation can be facilitated with concept distillation from generative image foundation models. 

\noindent\textbf{Acknowledgements:} The authors gratefully acknowledge the scientific support and HPC resources provided by the Erlangen National High Performance Computing Center (NHR@FAU) of the Friedrich-Alexander-Universität Erlangen-Nürnberg (FAU) under the NHR projects b143dc and b180dc. NHR funding is provided by federal and Bavarian state authorities. NHR@FAU hardware is partially funded by the German Research Foundation (DFG) – 440719683. Additional support was also received by the ERC - project MIA-NORMAL 101083647,  DFG KA 5801/2-1, INST 90/1351-1 and by the state of Bavarian.
H. Reynaud was supported by Ultromics Ltd. and the UKRI Centre or Doctoral Training in Artificial Intelligence for Healthcare (EP/S023283/1).

{\small
\bibliographystyle{ieee_fullname}
\bibliography{main}

\begin{thebibliography}{10}\itemsep=-1pt

\bibitem{abdal2021labels4free}
Rameen Abdal, Peihao Zhu, Niloy~J Mitra, and Peter Wonka.
\newblock Labels4free: Unsupervised segmentation using stylegan.
\newblock In {\em Proceedings of the IEEE/CVF International Conference on
  Computer Vision}, pages 13970--13979, 2021.

\bibitem{baranchuk2021label}
Dmitry Baranchuk, Andrey Voynov, Ivan Rubachev, Valentin Khrulkov, and Artem
  Babenko.
\newblock Label-efficient semantic segmentation with diffusion models.
\newblock In {\em International Conference on Learning Representations}, 2021.

\bibitem{benny2020onegan}
Yaniv Benny and Lior Wolf.
\newblock Onegan: Simultaneous unsupervised learning of conditional image
  generation, foreground segmentation, and fine-grained clustering.
\newblock In {\em European Conference on Computer Vision}, pages 514--530.
  Springer, 2020.

\bibitem{bielski2019emergence}
Adam Bielski and Paolo Favaro.
\newblock Emergence of object segmentation in perturbed generative models.
\newblock {\em Advances in Neural Information Processing Systems}, 32, 2019.

\bibitem{boecking2022making}
Benedikt Boecking, Naoto Usuyama, Shruthi Bannur, Daniel~C Castro, Anton
  Schwaighofer, Stephanie Hyland, Maria Wetscherek, Tristan Naumann, Aditya
  Nori, Javier Alvarez-Valle, et~al.
\newblock Making the most of text semantics to improve biomedical
  vision--language processing.
\newblock In {\em Computer Vision--ECCV 2022: 17th European Conference, Tel
  Aviv, Israel, October 23--27, 2022, Proceedings, Part XXXVI}, pages 1--21.
  Springer, 2022.

\bibitem{bommasani2021opportunities}
Rishi Bommasani, Drew~A Hudson, Ehsan Adeli, Russ Altman, Simran Arora, Sydney
  von Arx, Michael~S Bernstein, Jeannette Bohg, Antoine Bosselut, Emma
  Brunskill, et~al.
\newblock On the opportunities and risks of foundation models.
\newblock {\em arXiv:2108.07258}, 2021.

\bibitem{brown2020language}
Tom Brown, Benjamin Mann, Nick Ryder, Melanie Subbiah, Jared~D Kaplan, Prafulla
  Dhariwal, Arvind Neelakantan, Pranav Shyam, Girish Sastry, Amanda Askell,
  et~al.
\newblock Language models are few-shot learners.
\newblock {\em Advances in neural information processing systems},
  33:1877--1901, 2020.

\bibitem{chambon2022roentgen}
Pierre Chambon, Christian Bluethgen, Jean-Benoit Delbrouck, Rogier Van~der
  Sluijs, Ma{\l}gorzata Po{\l}acin, Juan Manuel~Zambrano Chaves, Tanishq~Mathew
  Abraham, Shivanshu Purohit, Curtis~P Langlotz, and Akshay Chaudhari.
\newblock Roentgen: Vision-language foundation model for chest x-ray
  generation.
\newblock {\em arXiv preprint arXiv:2211.12737}, 2022.

\bibitem{changpinyo2021conceptual}
Soravit Changpinyo, Piyush Sharma, Nan Ding, and Radu Soricut.
\newblock Conceptual 12m: Pushing web-scale image-text pre-training to
  recognize long-tail visual concepts.
\newblock In {\em CVPR'21}, pages 3558--3568, 2021.

\bibitem{chen2017deeplab}
Liang-Chieh Chen, George Papandreou, Iasonas Kokkinos, Kevin Murphy, and Alan~L
  Yuille.
\newblock Deeplab: Semantic image segmentation with deep convolutional nets,
  atrous convolution, and fully connected crfs.
\newblock {\em IEEE transactions on pattern analysis and machine intelligence},
  40(4):834--848, 2017.

\bibitem{chen2019unsupervised}
Micka{\"e}l Chen, Thierry Arti{\`e}res, and Ludovic Denoyer.
\newblock Unsupervised object segmentation by redrawing.
\newblock {\em Advances in neural information processing systems}, 32, 2019.

\bibitem{cheng2016long}
Jianpeng Cheng, Li Dong, and Mirella Lapata.
\newblock Long short-term memory-networks for machine reading.
\newblock In {\em Proceedings of the 2016 Conference on Empirical Methods in
  Natural Language Processing}, pages 551--561, 2016.

\bibitem{chung2022improving}
Hyungjin Chung, Byeongsu Sim, Dohoon Ryu, and Jong~Chul Ye.
\newblock Improving diffusion models for inverse problems using manifold
  constraints.
\newblock {\em arXiv preprint arXiv:2206.00941}, 2022.

\bibitem{chung2022come}
Hyungjin Chung, Byeongsu Sim, and Jong~Chul Ye.
\newblock Come-closer-diffuse-faster: Accelerating conditional diffusion models
  for inverse problems through stochastic contraction.
\newblock In {\em Proceedings of the IEEE/CVF Conference on Computer Vision and
  Pattern Recognition}, pages 12413--12422, 2022.

\bibitem{croitoru2022diffusion}
Florinel-Alin Croitoru, Vlad Hondru, Radu~Tudor Ionescu, and Mubarak Shah.
\newblock Diffusion models in vision: A survey.
\newblock {\em arXiv preprint arXiv:2209.04747}, 2022.

\bibitem{deng2009imagenet}
Jia Deng, Wei Dong, Richard Socher, Li-Jia Li, Kai Li, and Li Fei-Fei.
\newblock Imagenet: A large-scale hierarchical image database.
\newblock In {\em CVPR'09}, pages 248--255. Ieee, 2009.

\bibitem{devlin2018bert}
Jacob Devlin, Ming-Wei Chang, Kenton Lee, and Kristina Toutanova.
\newblock Bert: Pre-training of deep bidirectional transformers for language
  understanding.
\newblock {\em arXiv preprint arXiv:1810.04805}, 2018.

\bibitem{dhariwal2021diffusion}
Prafulla Dhariwal and Alexander Nichol.
\newblock Diffusion models beat gans on image synthesis.
\newblock {\em Advances in Neural Information Processing Systems},
  34:8780--8794, 2021.

\bibitem{gal2022image}
Rinon Gal, Yuval Alaluf, Yuval Atzmon, Or Patashnik, Amit~H Bermano, Gal
  Chechik, and Daniel Cohen-Or.
\newblock An image is worth one word: Personalizing text-to-image generation
  using textual inversion.
\newblock {\em arXiv preprint arXiv:2208.01618}, 2022.

\bibitem{galleguillos2010context}
Carolina Galleguillos and Serge Belongie.
\newblock Context based object categorization: A critical survey.
\newblock {\em Computer vision and image understanding}, 114(6):712--722, 2010.

\bibitem{hertz2022promptgoogleattentionediting}
Amir Hertz, Ron Mokady, Jay Tenenbaum, Kfir Aberman, Yael Pritch, and Daniel
  Cohen-Or.
\newblock Prompt-to-prompt image editing with cross attention control.
\newblock {\em arXiv preprint arXiv:2208.01626}, 2022.

\bibitem{ho2020denoisingddpm}
Jonathan Ho, Ajay Jain, and Pieter Abbeel.
\newblock Denoising diffusion probabilistic models.
\newblock {\em Advances in Neural Information Processing Systems},
  33:6840--6851, 2020.

\bibitem{ho2022classifier}
Jonathan Ho and Tim Salimans.
\newblock Classifier-free diffusion guidance.
\newblock {\em arXiv preprint arXiv:2207.12598}, 2022.

\bibitem{h36m_pami_human3point6}
Catalin Ionescu, Dragos Papava, Vlad Olaru, and Cristian Sminchisescu.
\newblock Human3.6m: Large scale datasets and predictive methods for 3d human
  sensing in natural environments.
\newblock {\em IEEE Transactions on Pattern Analysis and Machine Intelligence},
  36(7):1325--1339, jul 2014.

\bibitem{ji2019invariant}
Xu Ji, Joao~F Henriques, and Andrea Vedaldi.
\newblock Invariant information clustering for unsupervised image
  classification and segmentation.
\newblock In {\em Proceedings of the IEEE/CVF International Conference on
  Computer Vision}, pages 9865--9874, 2019.

\bibitem{Jiang_2022_CVPR}
Peng-Tao Jiang, Yuqi Yang, Qibin Hou, and Yunchao Wei.
\newblock L2g: A simple local-to-global knowledge transfer framework for weakly
  supervised semantic segmentation.
\newblock In {\em Proceedings of the IEEE/CVF Conference on Computer Vision and
  Pattern Recognition (CVPR)}, pages 16886--16896, June 2022.

\bibitem{johnson2019mimic}
Alistair~EW Johnson, Tom~J Pollard, Seth~J Berkowitz, Nathaniel~R Greenbaum,
  Matthew~P Lungren, Chih-ying Deng, Roger~G Mark, and Steven Horng.
\newblock Mimic-cxr, a de-identified publicly available database of chest
  radiographs with free-text reports.
\newblock {\em Scientific data}, 6(1):317, 2019.

\bibitem{KhoslaYaoJayadevaprakashFeiFei_FGVC2011}
Aditya Khosla, Nityananda Jayadevaprakash, Bangpeng Yao, and Li Fei-Fei.
\newblock Novel dataset for fine-grained image categorization.
\newblock In {\em First Workshop on Fine-Grained Visual Categorization, IEEE
  Conference on Computer Vision and Pattern Recognition}, Colorado Springs, CO,
  June 2011.

\bibitem{kim2020unsupervised}
Wonjik Kim, Asako Kanezaki, and Masayuki Tanaka.
\newblock Unsupervised learning of image segmentation based on differentiable
  feature clustering.
\newblock {\em IEEE Transactions on Image Processing}, 29:8055--8068, 2020.

\bibitem{KrauseStarkDengFei-Fei_3DRR2013_stanford_cars}
Jonathan Krause, Michael Stark, Jia Deng, and Li Fei-Fei.
\newblock 3d object representations for fine-grained categorization.
\newblock In {\em 4th International IEEE Workshop on 3D Representation and
  Recognition (3dRR-13)}, Sydney, Australia, 2013.

\bibitem{lee2021railroad}
Seungho Lee, Minhyun Lee, Jongwuk Lee, and Hyunjung Shim.
\newblock Railroad is not a train: Saliency as pseudo-pixel supervision for
  weakly supervised semantic segmentation.
\newblock In {\em Proceedings of the IEEE/CVF conference on computer vision and
  pattern recognition}, pages 5495--5505, 2021.

\bibitem{liu2022pseudo_plms}
Luping Liu, Yi Ren, Zhijie Lin, and Zhou Zhao.
\newblock Pseudo numerical methods for diffusion models on manifolds.
\newblock {\em arXiv preprint arXiv:2202.09778}, 2022.

\bibitem{long2015fully}
Jonathan Long, Evan Shelhamer, and Trevor Darrell.
\newblock Fully convolutional networks for semantic segmentation.
\newblock In {\em Proceedings of the IEEE conference on computer vision and
  pattern recognition}, pages 3431--3440, 2015.

\bibitem{lu2016learning}
Zhiwu Lu, Zhenyong Fu, Tao Xiang, Peng Han, Liwei Wang, and Xin Gao.
\newblock Learning from weak and noisy labels for semantic segmentation.
\newblock {\em IEEE transactions on pattern analysis and machine intelligence},
  39(3):486--500, 2016.

\bibitem{melas2021finding}
Luke Melas-Kyriazi, Christian Rupprecht, Iro Laina, and Andrea Vedaldi.
\newblock Finding an unsupervised image segmenter in each of your deep
  generative models.
\newblock {\em arXiv preprint arXiv:2105.08127}, 2021.

\bibitem{noh2015learning}
Hyeonwoo Noh, Seunghoon Hong, and Bohyung Han.
\newblock Learning deconvolution network for semantic segmentation.
\newblock In {\em Proceedings of the IEEE international conference on computer
  vision}, pages 1520--1528, 2015.

\bibitem{oktay2018attention}
Ozan Oktay, Jo Schlemper, Loic~Le Folgoc, Matthew Lee, Mattias Heinrich,
  Kazunari Misawa, Kensaku Mori, Steven McDonagh, Nils~Y Hammerla, Bernhard
  Kainz, et~al.
\newblock Attention u-net: Learning where to look for the pancreas.
\newblock {\em arXiv preprint arXiv:1804.03999}, 2018.

\bibitem{ouali2020autoregressive}
Yassine Ouali, C{\'e}line Hudelot, and Myriam Tami.
\newblock Autoregressive unsupervised image segmentation.
\newblock In {\em European Conference on Computer Vision}, pages 142--158.
  Springer, 2020.

\bibitem{parikhetal2016decomposable}
Ankur Parikh, Oscar T{\"a}ckstr{\"o}m, Dipanjan Das, and Jakob Uszkoreit.
\newblock A decomposable attention model for natural language inference.
\newblock In {\em Proceedings of the 2016 Conference on Empirical Methods in
  Natural Language Processing}, pages 2249--2255, Austin, Texas, Nov. 2016.
  Association for Computational Linguistics.

\bibitem{rajchl2016deepcut}
Martin Rajchl, Matthew~CH Lee, Ozan Oktay, Konstantinos Kamnitsas, Jonathan
  Passerat-Palmbach, Wenjia Bai, Mellisa Damodaram, Mary~A Rutherford, Joseph~V
  Hajnal, Bernhard Kainz, et~al.
\newblock Deepcut: Object segmentation from bounding box annotations using
  convolutional neural networks.
\newblock {\em IEEE transactions on medical imaging}, 36(2):674--683, 2016.

\bibitem{ramesh2022hierarchical}
Aditya Ramesh, Prafulla Dhariwal, Alex Nichol, Casey Chu, and Mark Chen.
\newblock Hierarchical text-conditional image generation with clip latents.
\newblock {\em arXiv preprint arXiv:2204.06125}, 2022.

\bibitem{ramesh2021zero}
Aditya Ramesh, Mikhail Pavlov, Gabriel Goh, Scott Gray, Chelsea Voss, Alec
  Radford, Mark Chen, and Ilya Sutskever.
\newblock Zero-shot text-to-image generation.
\newblock In {\em International Conference on Machine Learning}, pages
  8821--8831. PMLR, 2021.

\bibitem{remez2018learning}
Tal Remez, Jonathan Huang, and Matthew Brown.
\newblock Learning to segment via cut-and-paste.
\newblock In {\em Proceedings of the European conference on computer vision
  (ECCV)}, pages 37--52, 2018.

\bibitem{rombach2022high_latent_stable_diffusion}
Robin Rombach, Andreas Blattmann, Dominik Lorenz, Patrick Esser, and Bj{\"o}rn
  Ommer.
\newblock High-resolution image synthesis with latent diffusion models.
\newblock In {\em Proceedings of the IEEE/CVF Conference on Computer Vision and
  Pattern Recognition}, pages 10684--10695, 2022.

\bibitem{ronneberger2015u}
Olaf Ronneberger, Philipp Fischer, and Thomas Brox.
\newblock U-net: Convolutional networks for biomedical image segmentation.
\newblock In {\em International Conference on Medical image computing and
  computer-assisted intervention}, pages 234--241. Springer, 2015.

\bibitem{rother2004grabcut}
Carsten Rother, Vladimir Kolmogorov, and Andrew Blake.
\newblock ``grabcut'' interactive foreground extraction using iterated graph
  cuts.
\newblock {\em ACM transactions on graphics (TOG)}, 23(3):309--314, 2004.

\bibitem{ruiz2022dreambooth}
Nataniel Ruiz, Yuanzhen Li, Varun Jampani, Yael Pritch, Michael Rubinstein, and
  Kfir Aberman.
\newblock Dreambooth: Fine tuning text-to-image diffusion models for
  subject-driven generation.
\newblock {\em arXiv preprint arXiv:2208.12242}, 2022.

\bibitem{saharia2022photorealistic}
Chitwan Saharia, William Chan, Saurabh Saxena, Lala Li, Jay Whang, Emily
  Denton, Seyed Kamyar~Seyed Ghasemipour, Burcu~Karagol Ayan, S~Sara Mahdavi,
  Rapha~Gontijo Lopes, et~al.
\newblock Photorealistic text-to-image diffusion models with deep language
  understanding.
\newblock {\em arXiv preprint arXiv:2205.11487}, 2022.

\bibitem{savarese2021information}
Pedro Savarese, Sunnie~SY Kim, Michael Maire, Greg Shakhnarovich, and David
  McAllester.
\newblock Information-theoretic segmentation by inpainting error maximization.
\newblock In {\em Proceedings of the IEEE/CVF Conference on Computer Vision and
  Pattern Recognition}, pages 4029--4039, 2021.

\bibitem{schlemper2017deep}
Jo Schlemper, Jose Caballero, Joseph~V Hajnal, Anthony Price, and Daniel
  Rueckert.
\newblock A deep cascade of convolutional neural networks for mr image
  reconstruction.
\newblock In {\em IPMI'17}, pages 647--658. Springer, 2017.

\bibitem{sheikh2009background}
Yaser Sheikh, Omar Javed, and Takeo Kanade.
\newblock Background subtraction for freely moving cameras.
\newblock In {\em 2009 IEEE 12th International Conference on Computer Vision},
  pages 1219--1225. IEEE, 2009.

\bibitem{simonyan2014very}
Karen Simonyan and Andrew Zisserman.
\newblock Very deep convolutional networks for large-scale image recognition.
\newblock {\em arXiv:1409.1556}, 2014.

\bibitem{singer2022make}
Uriel Singer, Adam Polyak, Thomas Hayes, Xi Yin, Jie An, Songyang Zhang, Qiyuan
  Hu, Harry Yang, Oron Ashual, Oran Gafni, et~al.
\newblock Make-a-video: Text-to-video generation without text-video data.
\newblock {\em arXiv preprint arXiv:2209.14792}, 2022.

\bibitem{singh2019finegan}
Krishna~Kumar Singh, Utkarsh Ojha, and Yong~Jae Lee.
\newblock Finegan: Unsupervised hierarchical disentanglement for fine-grained
  object generation and discovery.
\newblock In {\em Proceedings of the IEEE/CVF conference on computer vision and
  pattern recognition}, pages 6490--6499, 2019.

\bibitem{song2020denoisingddim}
Jiaming Song, Chenlin Meng, and Stefano Ermon.
\newblock Denoising diffusion implicit models.
\newblock {\em arXiv preprint arXiv:2010.02502}, 2020.

\bibitem{tao2020hierarchical}
Andrew Tao, Karan Sapra, and Bryan Catanzaro.
\newblock Hierarchical multi-scale attention for semantic segmentation.
\newblock {\em arXiv preprint arXiv:2005.10821}, 2020.

\bibitem{vaswani2017attention}
Ashish Vaswani, Noam Shazeer, Niki Parmar, Jakob Uszkoreit, Llion Jones,
  Aidan~N Gomez, {\L}ukasz Kaiser, and Illia Polosukhin.
\newblock Attention is all you need.
\newblock {\em Advances in neural information processing systems}, 30, 2017.

\bibitem{WahCUB_200_2011}
C. Wah, S. Branson, P. Welinder, P. Perona, and S. Belongie.
\newblock The caltech-ucsd birds-200-2011 dataset.
\newblock Technical Report CNS-TR-2011-001, California Institute of Technology,
  2011.

\bibitem{wang2022pretraining}
Tengfei Wang, Ting Zhang, Bo Zhang, Hao Ouyang, Dong Chen, Qifeng Chen, and
  Fang Wen.
\newblock Pretraining is all you need for image-to-image translation.
\newblock {\em arXiv preprint arXiv:2205.12952}, 2022.

\bibitem{yang2022diffusion}
Ling Yang, Zhilong Zhang, Yang Song, Shenda Hong, Runsheng Xu, Yue Zhao,
  Yingxia Shao, Wentao Zhang, Bin Cui, and Ming-Hsuan Yang.
\newblock Diffusion models: A comprehensive survey of methods and applications.
\newblock {\em arXiv preprint arXiv:2209.00796}, 2022.

\bibitem{yang2022learningforeground_background_segm}
Yu Yang, Hakan Bilen, Qiran Zou, Wing~Yin Cheung, and Xiangyang Ji.
\newblock Learning foreground-background segmentation from improved layered
  gans.
\newblock In {\em Proceedings of the IEEE/CVF Winter Conference on Applications
  of Computer Vision}, pages 2524--2533, 2022.

\bibitem{yuan2020object}
Yuhui Yuan, Xilin Chen, and Jingdong Wang.
\newblock Object-contextual representations for semantic segmentation.
\newblock In {\em European conference on computer vision}, pages 173--190.
  Springer, 2020.

\bibitem{zhong2020squeeze}
Zilong Zhong, Zhong~Qiu Lin, Rene Bidart, Xiaodan Hu, Ibrahim~Ben Daya, Zhifeng
  Li, Wei-Shi Zheng, Jonathan Li, and Alexander Wong.
\newblock Squeeze-and-attention networks for semantic segmentation.
\newblock In {\em Proceedings of the IEEE/CVF conference on computer vision and
  pattern recognition}, pages 13065--13074, 2020.

\end{thebibliography}
}

\clearpage
\appendix

\section{Evaluation Strategy}
We evaluate the accuracy of our masks by training a plain U-Net on the task of binary classification, following an approach close to the one proposed by \cite{yang2022learningforeground_background_segm}. 
For CUB we use the provided segmentation masks as ground-truth. 
For all the other datasets, we use the provided bounding boxes. 
We train the U-Net for 12,000 steps using a batch size of 32 and Adam optimizer with a learning rate 0.001. 
During training, we crop images randomly to $128 \times 128$ pixels and during inference, we employ center-cropping.   

\section{Finetuning}
\begin{figure}[h!]
\centering
\includegraphics[width=\linewidth]{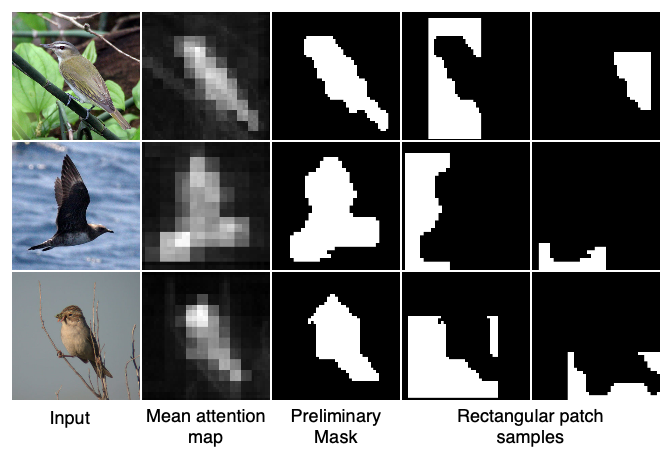}
\caption{Illustration of how we extract rectangular patches for background inpainting during finetuning.}
\label{fig:background_rect_masks}
\end{figure}

We fine-tune the diffusion models on the datasets by taking the  avenue provided by \cite{rombach2022high_latent_stable_diffusion}. 
Fine-tuning for foreground generation is straightforward by training the model to perform full image synthesis.
For background generation, we select a random rectangular patch from the image, exclude any pixels covered by the preliminary mask, and train the model to reproduce the remaining background pixels in the background (see \cref{fig:background_rect_masks} for examples).
The diffusion models are trained on foreground and background generation simultaneously, with each objective being trained in an equal proportion. 

Computation of the refined masks, which requires the computation of the preliminary masks, takes $7.5$ seconds for a batch of three samples on a single GPU.  

\section{Empirical Proof of Simplified Equation}

In this section, we show empirically that it is not necessary to repeat single diffusion steps in order to achieve better results on preliminary masks. 
Formally we evaluate  
\begin{equation}
\label{eq:eq_last_diff_step_simp}
    \hat{M} = \sum_{t=1}^{T_0}\mathbb{E}_{\mathbf{z} \sim p(\mathbf{z_t}|\mathbf{z_{t+1}}, \mathbf{z_0})}
    [
    \sum_{l} \psi_{\mathbf{z_{t,l}}}(Q_l, K^T_l)
    ], 
\end{equation}

and show that it can be simplified to 

\begin{equation}
    \label{eq:multiple_diff_steps_simp}
    \hat{M} = \sum_{t=1}^{T_0}\sum_{l}{\psi_{\mathbf{z_{t,l}}}(Q_l, K_l^T)}
\end{equation}

by computing accuracy metrics on CUB for the case of $T_0 = 1$. 

\begin{figure}[h!]
    \centering
    \includegraphics[width=\linewidth]{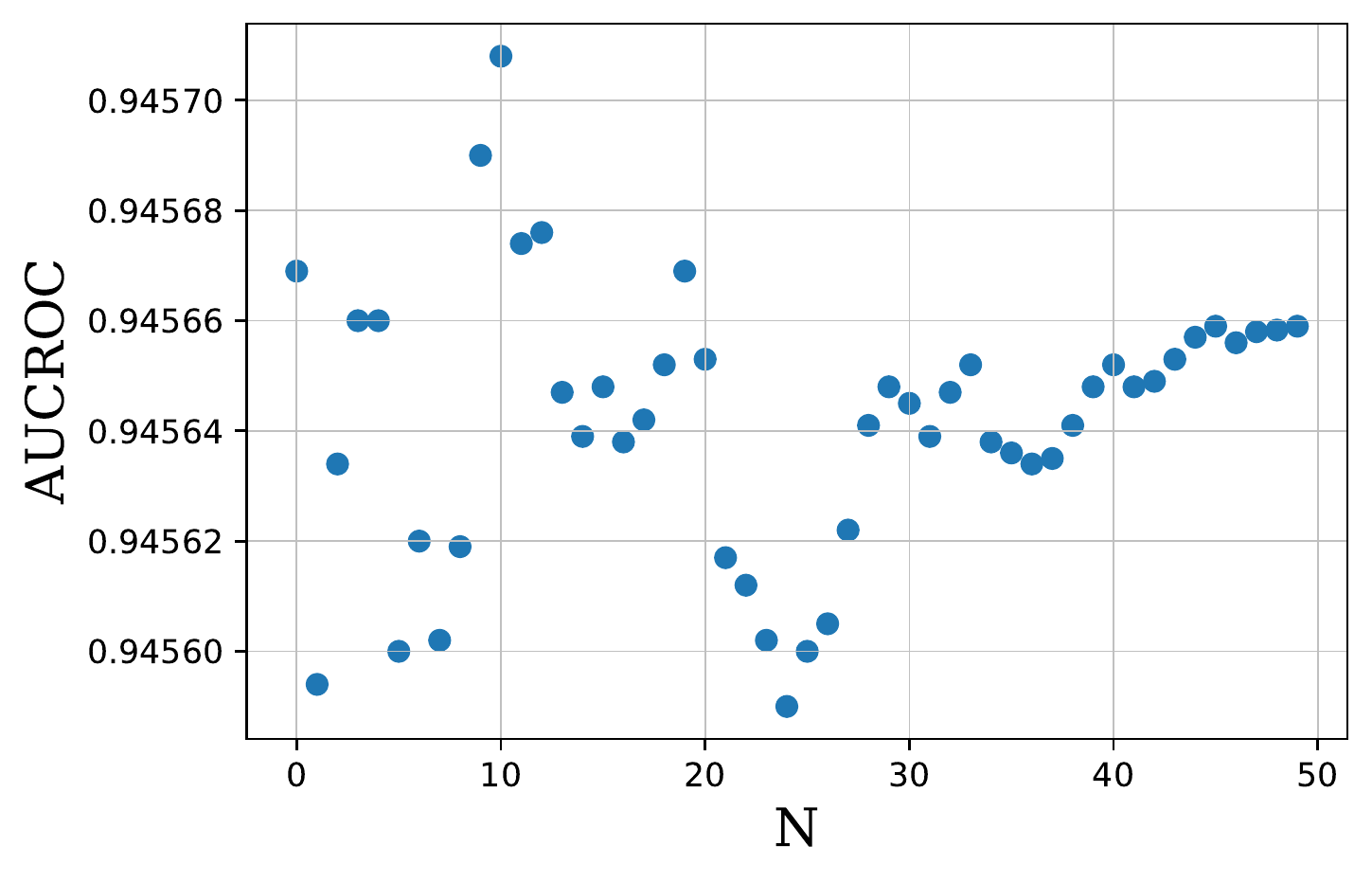}
    \caption{AUCROC of CUB over an increasing number of monte carlo samples $N$. }
    \label{fig:expectation_over_final_diff_steps}
\end{figure}

The motivation behind this is that the computation of the preliminary masks is the bottleneck of our pipeline, and we want the computation time to remain reasonable. The execution time increases linearly with the number of repetitions. We estimate the expectation in \Cref{eq:eq_last_diff_step_simp} using Monte Carlo sampling and denote the number of samples as $N$ and compute the AUCROC as a function over it.
The results are shown in \cref{fig:expectation_over_final_diff_steps}. These results suggest that increasing $N$ also slightly increases the absolute AUCROC value, while simultaneously decreasing the variance. However, these improvements are within a very small margin. 
Intuitively this means that the diffusion model is quite robust towards different latent inputs $z_t$. We conclude from this that it is unreasonable to compute attention masks over multiple steps and therefore perform all experiments using \cref{eq:multiple_diff_steps_simp}.  


\begin{figure}[h!]
\centering
\includegraphics[width=\linewidth]{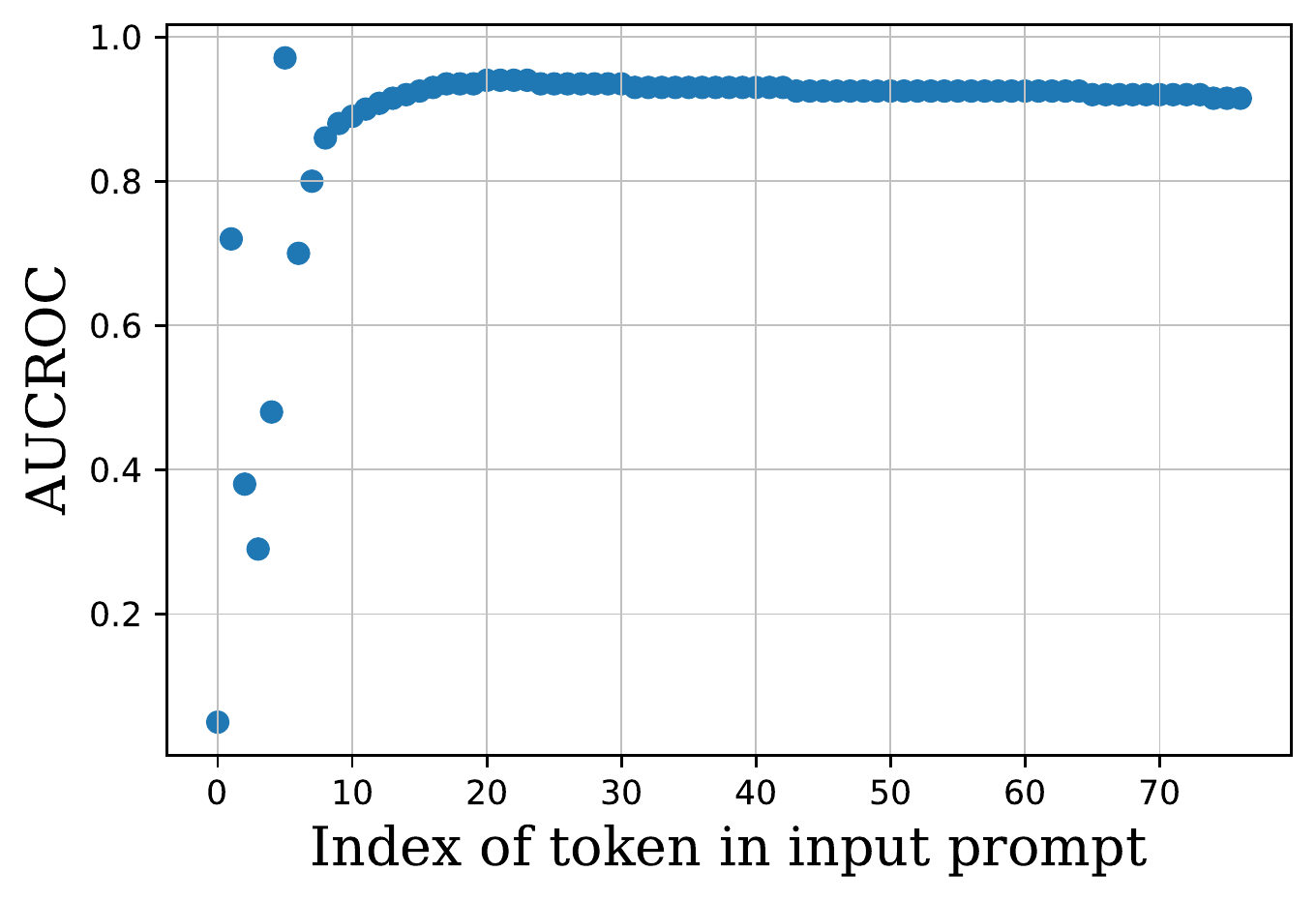}
\caption{AUCROC values of the preliminary masks extracted for every token using the prompt ``a photo of a bird". 
The language model is inherited from \cite{rombach2022high_latent_stable_diffusion} and uses a BERT-tokenizer \cite{devlin2018bert}. 
Therefore a ``startofstring" token is added to the beginning of the input and the length is padded to a length of 77. The peak is at the ``bird" token.}
\label{fig:roc_prompt_engineering}
\end{figure}
Initially, we also experimented using only the object token or using a list of objects as an input prompt but both approaches result in more noisy preliminary masks and inferior AUCROC values.  

\section{Outlier Cases}
\begin{figure}[h!]
\centering
\includegraphics[width=\linewidth]{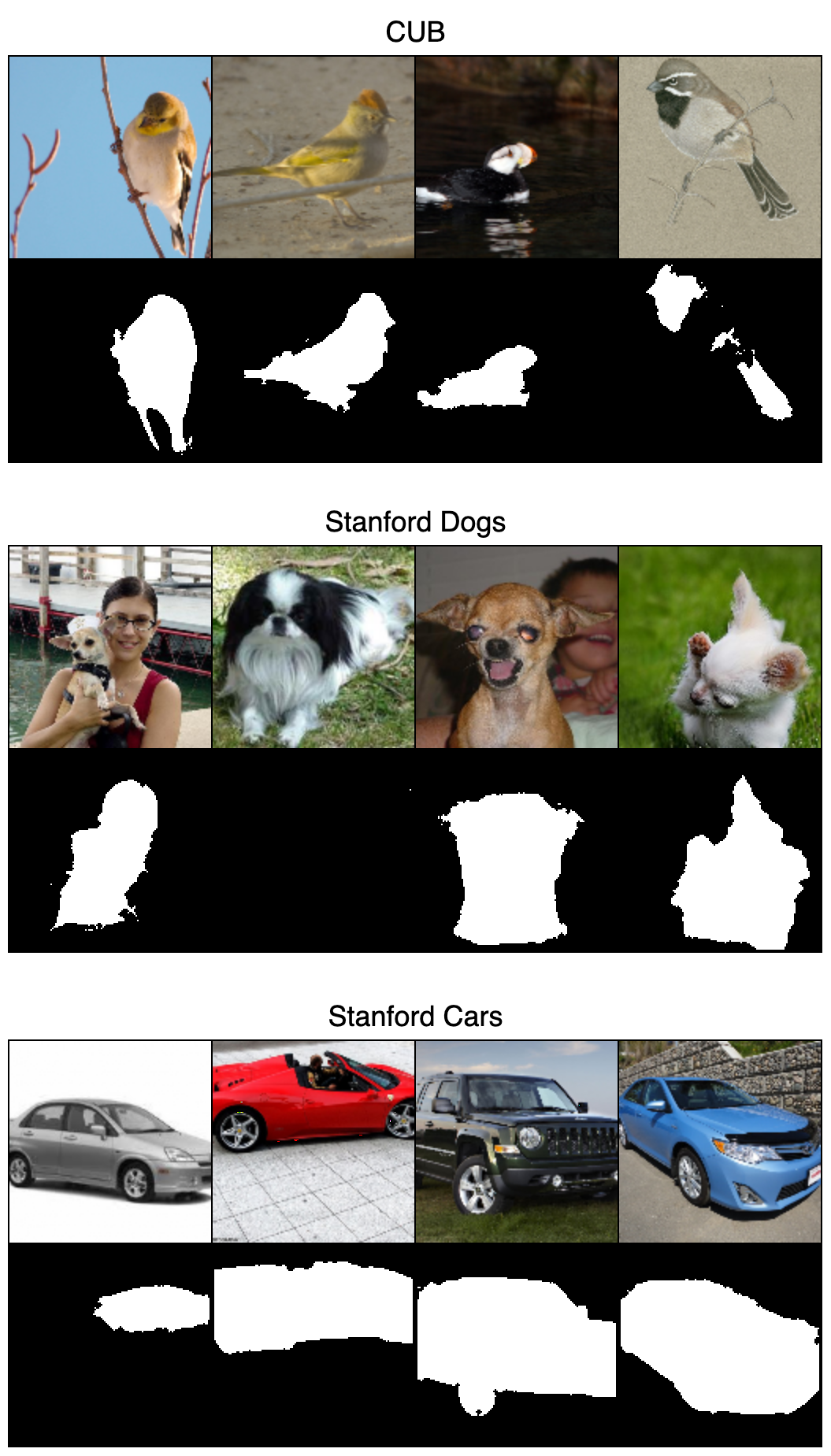}
\caption{Examples of failure cases of the U-Net model.}
\label{fig:failure_cases}
\end{figure}

Since our method is self-supervised it can be prone to  some errors, as the model has never seen an example of a real segmentation.
In Fig \ref{fig:failure_cases} we show examples of rare selected  failure cases that we have observed during testing.

The first is that branches on which birds are sitting are often segmented as part of the bird.
This is a result of the preliminary masks which sometimes include these branches into objects. Consequently, the U-Net is uncertain about these areas.
We observe that this only happens to birds that are clinging to branches, as can be seen in the left image in \cref{fig:failure_cases}.
In rare occasions, very low contrast also leads the model to accidentally predict parts of the image as background (such as in the right two bird examples).
This is likely because the pixel intensity distribution of the bird is too close to that of the background, causing the image difference when computing the refined masks to be too small, resulting in it being misclassified as background.

In the dog dataset, the method often struggled if humans were holding the dog (see left-most dog example).
In these cases, the final segmentation only excludes parts of the human in the background but not all of it.
One possible cause of this is that the model struggles to reconstruct the human when performing background inpainting, due to feature complexity, causing it to have a high-intensity difference when computing the refined masks.
We observed that this did not happen if the humans were positioned further back in the background, as can be seen in the third example in Figure~\ref{fig:failure_cases}. 
Finally, the method also struggled with dogs that are only black and white.
We believe that this is because of a bimodal pixel distribution assumption. 
If we perform inpainting for mask refinement for these dogs the white and black parts have very different contrasts compared to the background and are consequently assigned different modes of the bimodal Gaussian mixture model.  

In the case of the cars our method seemed to have limited performance if the input image had a plain white background. 
We believe this is because during the refinement stage the model does not expect the image to be entirely empty, and therefore always tries to inpaint something in the image center. 
However, the problem of segmenting cars in front of white background can be solved using trivial methods.
The predictions for Human3.6m were consistent throughout the whole dataset, with the exception of occasional under-segmentation of the legs, as explained in the main paper. 

\section{Inpainting Ablation}
To verify that our inpainting strategy does not make our proposed pipeline unnecessarily complicated, we try to refine the masks using a simpler approach that crops regions of the background and uses them to inpaint. 
We do this by extracting the largest background region according to the preliminary masks and then flipping it into the region of the foreground object.

\section{Prompt Engineering}
\begin{figure}
    \centering
    \includegraphics[width=\linewidth]{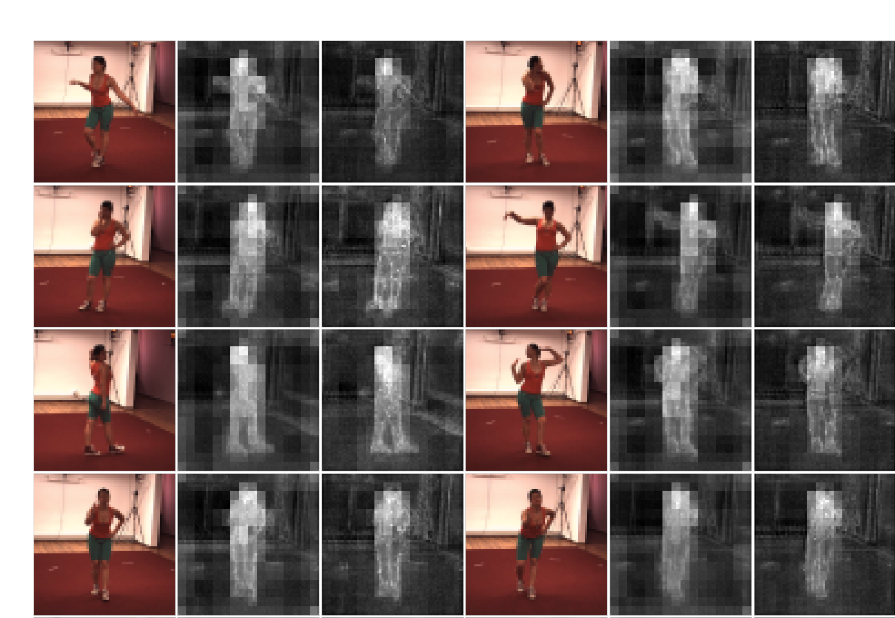}
    \caption{Preliminary masks computed on the prompt 
    ``A photo of a human standing in a room'' (left) and on the prompt ``A photo of a human'' (right).}
    \label{fig:prompt_engineering_with_background}
\end{figure}

To further justify the choice of our text-conditioning $y$ we compute the AUCROC for every token of the prompt ``a photo of a bird". 
Internally this prompt is preceded by a fixed ``startofstring" token. 
The results are shown in \cref{fig:roc_prompt_engineering}. 
We can clearly observe that the highest response happens if we compute $\hat{M}$ for the token ``bird".
After that, the AUCROC remains high, albeit not at the same level as before. 
We decide against incorporating the preliminary maps of different tokens into the pipeline because they slightly decrease the AUCROC value for the preliminary masks while simultaneously reducing the interpretability of our approach. 
The same holds true for the ``startofstring" token, which has very low activation on all the bird pixels. By inverting the attention scores we could therefore also locate objects.
However, this observation is a direct consequence of the computation of the attention probabilities. 
Attention is computed for every pixel as a probability of belonging to a token using softmax normalization on the attention scores. 
Since this probability has to sum up to one, the activations of non-object pixels have to be high for some tokens.
Furthermore, we analyzed the stability of the extracted preliminary masks in terms of minor changes to the input prompt. 
In \cref{fig:prompt_engineering_with_background}, we show the difference of the segmentation masks if we integrate more prior knowledge by describing the image composition in the prompt. From these images, we can see that there are only minor changes to the silhouette of the human and, consequently, that the results are mostly independent of the prompt.
We made the same observation when prompting on ``person'' instead.

\section{Classifier-free Guidance}
We use classifier-free guidance as proposed by~\cite{ho2022classifier}:

\begin{equation}
    \tilde{\epsilon}_\theta(\mathbf{z}_t, y_f) 
    = w\mathbf{\epsilon}_\theta(\mathbf{z}_t, y_f) 
       - (w - 1) \mathbf{\epsilon}_\theta(\mathbf{z}_t, y),
\end{equation}
where $w$ denotes the classifier-free guidance scale, and $\epsilon$ the update term of the diffusion process.   
In our case, we assume that the latent space representation of the images $\mathbf{z}_t$ conditioned on the prompts is reduced to the background and the foreground clusters. Consequently, we can replace the unconditional prompt with the background prompt from the equation, which changes it to
\begin{equation}
    \tilde{\epsilon}_\theta(\mathbf{z}_t, y_f) 
    = w\mathbf{\epsilon}_\theta(\mathbf{z}_t, y_f) 
       - (w - 1) \mathbf{\epsilon}_\theta(\mathbf{z}_t, y_b).
\end{equation}

Finally, we can also perform classifier-free guidance for background generation by setting the scale to $w=-1$ which is equivalent to switching the prompts and setting $w=2$. 

\begin{figure*}
    \centering
    \includegraphics[width=\linewidth]{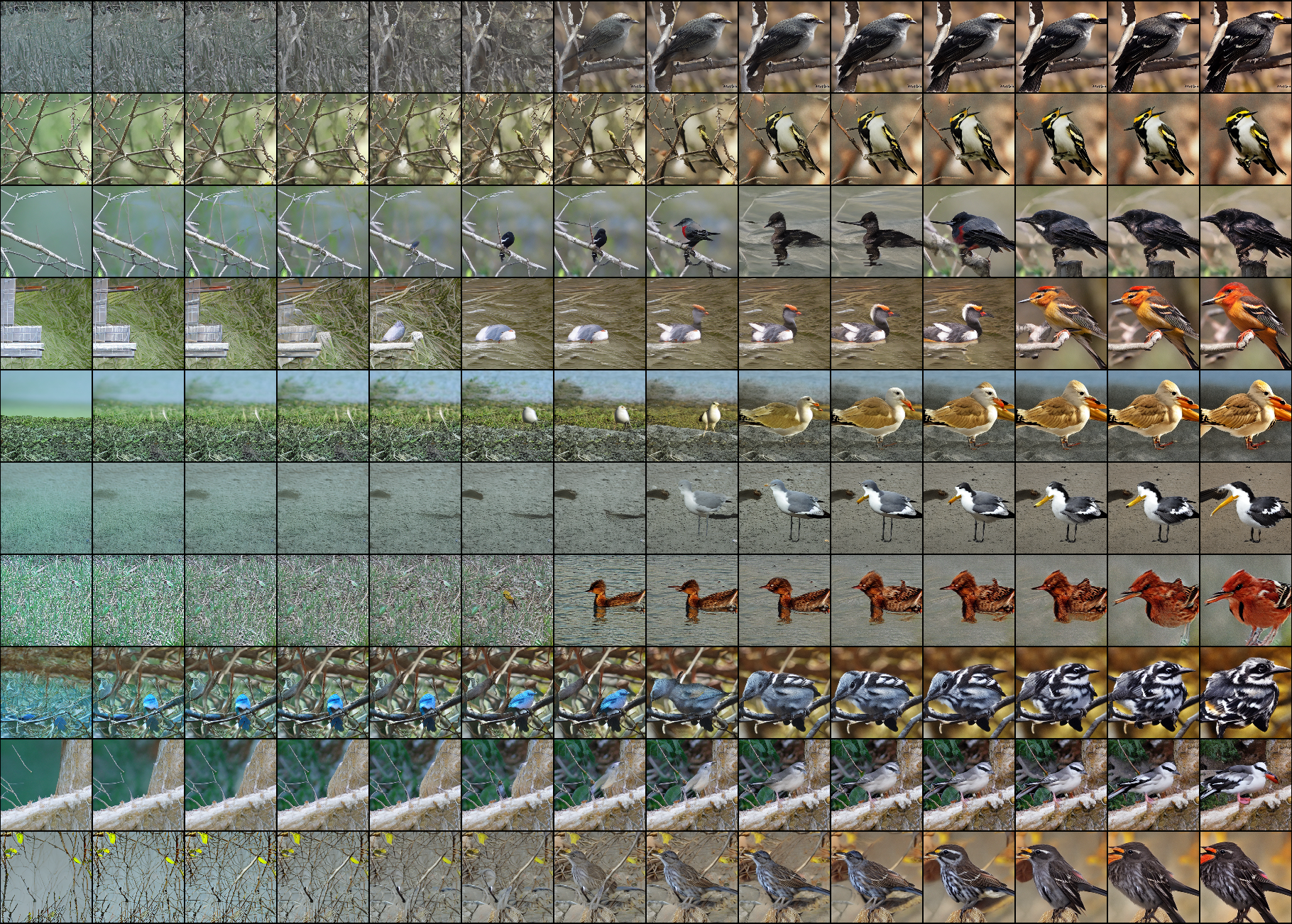}
    \caption{Synthesis results starting from the same seeds while increasing the scale of classifier-free guidance. The guidance scale $w$ ranges from -6.5 to 6.5 and is increased in steps of 1.}
    \label{fig:cfg_grid_full}
\end{figure*}

To verify this, \cref{fig:cfg_grid_full} shows the influence of $w$ in more detail. 
Images with high guidance towards the background (\emph{i.e.}, low $w$) do not show any signs of the object. By increasing this value, we can see a bird growing from a part of the image.  
To further illustrate this process we added a few video samples of this to the supplements. 
Judging from these images we concluded that our assumption of the clustering is correct and that the model has indeed learned what background information is. 


\section{Medical Image Analysis}
To analyze whether LDMs are interpretable after being adapted to domain-specific tasks, we evaluate our proposed extraction method on an LDM fine-tuned on MIMIC \cite{johnson2019mimic} following an approach similar to the one suggested by \cite{chambon2022roentgen}. 
Fine-tuning is done for 60k steps over $\sim$160000 images and the \textit{impression} section of the radiology reports corresponding to the images. The learning rate is set to $5\times 10^{-5}$, and the language encoder is kept frozen. 
We set the batch size during fine-tuning to 256, spread over 16 80GB A100 GPUs during roughly 470 hours of computation. 
To evaluate the localization accuracy, we take the impressions of the MS-CXR subset \cite{boecking2022making}, which we left as a hold-out set during training.
Then, we use the impressions from \cite{boecking2022making} and compute $\hat{M}$ and $M_{pre}$ on the tokens corresponding to the eight different diseases of the dataset, and compare the predicted region with the ground-truth bounding boxes.
Because some words are unknown to the language encoder, they were split into different tokens. In this case, we compute the sum over the attention maps of all tokens.

\clearpage

\end{document}